\documentclass[twocolumn]{article}

\usepackage[T1]{fontenc}
\usepackage[utf8]{inputenc}
\usepackage{textcomp}
\usepackage{graphicx}
\usepackage{url}
\usepackage{amsmath}
\usepackage{amssymb}
\usepackage{amsfonts}
\usepackage{xcolor}
\usepackage{array}
\usepackage{float}
\usepackage{subcaption}
\usepackage{caption}
\usepackage{booktabs}
\usepackage{geometry}
\usepackage{multirow}
\usepackage{placeins}
\usepackage{hyperref}
\usepackage{authblk}
\usepackage{enumitem}
\usepackage{algorithm}
\usepackage{algpseudocode}
\usepackage{tikz}
\usetikzlibrary{shapes.geometric,arrows.meta,positioning,fit,backgrounds,calc}

\title{Fine-Tuning Vision-Language Models for Understanding Current Damage and Scoring Priority with Quality Guard Agent}

\author[1]{Takato Yasuno}
\date{}

\begin{document}

\renewcommand{\abstractname}{Abstract}
\renewcommand{\refname}{References}
\renewcommand{\figurename}{Figure}
\renewcommand{\tablename}{Table}

\maketitle

%% ====================================================================
\begin{abstract}
%% ====================================================================

Bridge inspection in Japan requires mandatory visual assessments every five years,
yet qualitative damage ratings (levels a--e) assigned by different engineers exhibit
significant inter-rater variability---a critical barrier to consistent infrastructure
management. The aging of skilled engineers further threatens inspection capacity.
This paper presents a methodology for automating bridge damage understanding and
repair priority scoring using fine-tuned Vision-Language Models (VLMs).

We fine-tune LLaVA-1.5-7B with QLoRA (4-bit NF4 quantization, rank=32) on
up to 4{,}000 paired bridge damage images and inspection text records, then
evaluate on a fixed test set of 800 images. The model outputs natural language
descriptions identifying structural members and damage patterns, from which
a rule-based scoring engine calculates a five-level repair priority index.
A progressive training study (1k/2k/3k/4k samples) reveals that 2k training
samples achieve near-optimal validation loss (3.073) in only 2.9 hours of
training; beyond 2k, validation loss improves by no more than 0.2\% per
doubling of training samples, exhibiting clear diminishing returns.
Furthermore, semantic similarity on the held-out test set peaks at 3k (0.6909)
and degrades at 4k (0.6739), indicating that quality-curated mid-scale data
outperforms larger but noisier corpora.
Inference optimization combining \texttt{torch.compile()} and batch processing
(batch\_size=8) achieves 10.06 seconds per image---a 70.2\% reduction over the
unoptimized baseline.

Our approach contributes to data governance in bridge inspection, reduces
inter-rater variability, and provides AI-assisted triage to augment
expert engineers in inspection workflows.
Furthermore, we introduce a two-stage \textbf{Quality Guard} using a
fine-tuned Swallow-8B SLM to reject low-quality VLM outputs before
priority scoring, preventing spurious scores from damaged or unrecognised images.

\end{abstract}


%% ====================================================================
\section{Introduction}
\label{sec:introduction}
%% ====================================================================

Japan's road network encompasses approximately 730{,}000 highway bridges;
nearly half of these will exceed 50 years of service life by
2033~\cite{mlit2023bridge}.
Under the Road Act (amended 2014), every road bridge must undergo close-up
visual inspection every five years, producing paired records: a photographic
image of the damage site and a hand-written inspection note (\textit{shoken})
describing the structural member, damage type, severity, and recommended
corrective action.

Despite this systematic framework, two persistent challenges impede inspection
quality.
\textbf{First, inter-rater variability:} damage ratings (levels a--e) assigned
by different engineers diverge due to subjective judgment, undermining data
consistency across the national bridge stock~\cite{mlit2023bridge}.
\textbf{Second, workforce contraction:} experienced bridge inspectors are
retiring faster than they can be replaced, threatening capacity at the
required cadence.

Recent Vision-Language Models (VLMs) present a promising path forward.
Models such as LLaVA~\cite{liu2024llava},
InstructBLIP~\cite{dai2023instructblip}, and GPT-4V~\cite{openai2023gpt4}
can generate detailed descriptive text from images; however, these
general-purpose models have no exposure to specialised Japanese inspection
vocabulary and cannot produce the structured output required by inspection
standards.

This paper presents a methodology for \textbf{bridge damage understanding and
repair priority scoring} via domain-adapted VLMs.
We fine-tune LLaVA-1.5-7B~\cite{liu2024llava} with
QLoRA~\cite{dettmers2023qlora} on up to 4{,}000 paired bridge inspection
records.
We then deploy a rule-based scoring engine that converts the model's
free-form descriptions into a five-level repair priority index
(Immediate to Minimal).

\textbf{Terminology and scope.}
We distinguish \textit{damage understanding} from \textit{damage description}.
\textit{Damage understanding} refers to the cognitive process by which an
engineer interprets the condition of a structure along four complementary
aspects: (1)~the \textbf{current state} of surface damage visible in an
image, (2)~the \textbf{temporal progression} of deterioration across
multiple inspection time points, (3)~the \textbf{latent causal factors}
(environmental exposure, traffic load, material aging) inferred from
domain knowledge and experience, and (4)~the \textbf{heterogeneity of
deterioration rates}---identifying accelerating versus stable clusters of
structures or members.
\textit{Damage description}, by contrast, denotes the natural-language text
produced as VLM inference output that records observable damage signals
from an input image.
This paper scopes the VLM-based methodology to aspect~(1) only: damage
understanding from a single current-time image, realised as a damage
description that feeds the priority-scoring rule engine.
Aspects~(2)--(4) require longitudinal data, contextual metadata, and
structure-level aggregation, and are positioned as future extensions in
Section~\ref{sec:disc_panel}.

\textbf{Contributions:}
\begin{enumerate}[leftmargin=*,itemsep=1pt]
  \item \textbf{Damage VLM}: the first QLoRA fine-tuning study of a
        7B-parameter VLM on Japanese bridge inspection image--text pairs,
        achieving the \textit{Acceptable} quality tier on a consumer GPU.
  \item \textbf{Repair Priority Scoring}: a transparent, auditable rule engine
        that converts VLM free-form output into a five-level priority index,
        enabling consistent maintenance triage without subjective judgment.
  \item \textbf{Quality Guard Agent}: a two-stage guard (rule-based Stage~1
        + Swallow-8B SLM judge Stage~2) that rejects low-quality outputs
        before scoring, with thresholds calibrated from the empirical
        5th/95th token-count percentiles of $n=800$ predictions (12.5\%
        low-quality rate identified).
  \item \textbf{Visual Inspection ScoreBot}: an end-to-end local pipeline
        combining the Damage VLM with the Quality Guard and scoring engine---deployable without
        cloud API dependencies on a single consumer GPU.
\end{enumerate}

This work builds on the authors' prior study of quantized LLaVA-1.5-7B for
bridge damage assessment~\cite{yasuno2026quantized}, which benchmarked
INT4/INT8/BF16 quantization levels without task-specific fine-tuning.
The multi-stage inspection framework in~\cite{yasuno2026multistage} further
motivated the end-to-end pipeline presented here.

Section~\ref{sec:related} reviews related work, including
Section~\ref{sec:related:guard} on AI agent and quality guardrail methods.
Section~\ref{sec:methodology} details the system methodology.
Section~\ref{sec:results} presents evaluation results.
Section~\ref{sec:discussion} discusses findings and limitations.
Section~\ref{sec:conclusion} concludes.

%% ====================================================================
\section{Related Work}
\label{sec:related}
%% ====================================================================

\subsection{Vision-Language Models for Structural Damage Understanding}

The development of large VLMs accelerated with CLIP~\cite{radford2021clip},
which established scalable joint image--text representation via contrastive
pre-training on 400 million web image--caption pairs.
Flamingo~\cite{alayrac2022flamingo} extended this to few-shot visual reasoning
by conditioning frozen language models on visual features through gated
cross-attention layers.
BLIP-2~\cite{li2023blip2} and InstructBLIP~\cite{dai2023instructblip} improved
instruction-following by bridging frozen image encoders and language models
via a lightweight querying transformer.
LLaVA~\cite{liu2024llava} simplified the paradigm by projecting visual tokens
directly into a language model embedding space, enabling visual instruction
tuning at reduced cost.
Its successor LLaVA-1.5~\cite{liu2024llava15}, a CVPR~2024 highlight paper,
showed that an MLP connector with CLIP-ViT-L-336\,px achieves state-of-the-art
across 11 benchmarks using only 1.2\,M publicly available training samples
--- the same base model adopted in this work.
MiniGPT-4~\cite{zhu2024minigpt4} demonstrated that aligning a frozen visual
encoder with a large language model through a single projection layer suffices
for rich image-description generation.

Foundational perception models have further expanded the toolkit for inspection.
Segment Anything (SAM)~\cite{kirillov2023sam} introduced a promptable model
trained on over one billion masks that transfers zero-shot across diverse
visual domains.
Grounding DINO~\cite{liu2023groundingdino} combines transformer-based detection
with language grounding for open-set localisation, complementary to the
language-generation focus of generative VLMs.

In civil infrastructure, vision-based inspection has followed an independent
trajectory focused on pixel-level detection.
Cha et al.~\cite{cha2017deepcrack} demonstrated that convolutional neural
networks detect surface cracks with near-expert accuracy.
Spencer et al.~\cite{spencer2019vision} surveyed computer vision advances for
structural inspection, noting the gap between pixel-level detection and the
natural language reporting required in engineering practice.
Dorafshan et al.~\cite{dorafshan2018concrete} benchmarked CNN detectors for
concrete crack localisation, and Hsieh and Tsai~\cite{hsieh2020concrete}
compared machine learning approaches for crack detection.

Our work bridges these two streams: task-specific QLoRA fine-tuning of
LLaVA-1.5 generates the structured natural language output required by
Japanese inspection standards, going beyond pixel-level detection to holistic
damage description in domain-appropriate Japanese vocabulary.
Early work on few-shot damage vision mining~\cite{yasuno2023fewshot}
showed that imbalanced data and rare damage events are fundamental
challenges in infrastructure inspection datasets.
A multi-stage bridge inspection system~\cite{yasuno2026multistage} integrated
foundation models with location anonymisation for privacy-preserving
field imaging.
Our quantization study~\cite{yasuno2026quantized} benchmarked inference
speed and description quality across LLaVA-1.5-7B precision levels,
establishing INT4 QLoRA as the optimal speed--quality trade-off
and motivating the domain fine-tuning step presented in this paper.

\subsection{AI-Assisted Decision Support for Infrastructure Maintenance}

Bridge management has long relied on formal decision-support frameworks.
Frangopol~\cite{frangopol2011lifecycle} formalised life-cycle performance
optimisation for structural systems under uncertainty, establishing the
theoretical basis for risk-based inspection prioritisation.
Operational bridge management systems encode engineering heuristics into
rule-based priority scores but depend on consistent structured input that
manual inspection cannot always guarantee.

Yang et al.~\cite{yang2018dam} applied deep CNNs to automatically map surface
cracks on dam structures, reducing manual effort in routine assessments.
Such detection-focused systems, however, lack the semantic understanding
needed to recommend maintenance actions or to assign priority across
heterogeneous damage types.

More recently, vision-language models have been adapted for zero-shot
industrial inspection.
WinCLIP~\cite{jeong2023winclip} (CVPR~2023) introduced a compositional
ensemble on window/patch/image-level CLIP features, achieving 91.8\% AUROC
in zero-shot anomaly classification on MVTec-AD without task-specific training.
AnomalyCLIP~\cite{zhou2024anomalyclip} (ICLR~2024) extended this line by
learning object-agnostic text prompts that capture generic normality and
abnormality, enabling strong zero-shot transfer across 17 datasets spanning
defect inspection and medical imaging.
AnomalyGPT~\cite{gu2024anomalygpt} (AAAI~2024) demonstrated that large
vision-language models fine-tuned with simulated anomalous images and
corresponding textual descriptions achieve state-of-the-art industrial
anomaly detection without manual threshold setting, reaching 94.1\% image-level
AUC with only one normal reference shot.

Our scoring engine addresses the bridge-specific gap: rather than detecting
anomalies from scratch, it consumes VLM-generated Japanese damage descriptions
and applies YAML-defined weights for member type, severity, location, and
extent to produce a five-level repair priority index.
Domain experts can audit and update the scoring logic without machine learning
expertise, preserving transparency for safety-critical decisions.
Complementary work on heterogeneous graph importance scoring for bridge
networks~\cite{yasuno2026graph} demonstrated that LLM-generated
interpretations effectively bridge algorithmic rankings and maintenance
decision-makers, reinforcing the value of language as an interface between
AI outputs and engineering judgment.

\subsection{Multimodal Systems for Industrial Visual Inspection}

The emergence of GPT-4V~\cite{openai2023gpt4} catalysed interest in
general-purpose multimodal systems for industrial applications.
Cao et al.~\cite{cao2023gpt4vanomaly} systematically evaluated GPT-4V on
anomaly detection across four modalities and nine tasks, demonstrating
promising zero-shot and one-shot performance while highlighting limitations
in fine-grained localisation of subtle defects.
Prompt-engineered GPT-4V can identify manufacturing defects and answer
technical questions from product photographs, yet proprietary API dependence,
latency, and data privacy constraints limit deployment in regulated
infrastructure contexts.

Open-source VLMs fine-tuned or adapted for domain-specific inspection offer
a more practical path.
Anomaly-OV~\cite{xu2025anomalyov} (CVPR~2025) proposed a specialist visual
assistant for zero-shot anomaly detection and reasoning, introducing a
Look-Twice Feature Matching mechanism that adaptively emphasises abnormal
visual tokens, achieving significant improvements over generalist MLLMs on
industrial and medical benchmarks.
Zhang et al.~\cite{zhang2025trainingfree} (CVPR~2025) demonstrated
training-free anomaly detection using vision and language foundation models,
eliminating fine-tuning while maintaining competitive performance on standard
benchmarks --- contrasting with our QLoRA-based domain adaptation strategy.
LogicAD~\cite{jin2025logicad} (AAAI~2025) combined VLM-based text feature
extraction with logical reasoning to produce explainable anomaly reports,
aligning with the engineering interpretability requirements of infrastructure
operators.

Our \textbf{Visual Inspection ScoreBot} operationalises this paradigm for
Japanese bridge infrastructure: the Damage VLM generates free-form Japanese
damage descriptions; a regex-based extractor converts them to structured
JSON attributes; and the scoring engine produces actionable priority
scores --- all running locally on a single consumer GPU without external API
calls.
This architecture aligns with emerging industrial AI patterns in which a
language model serves as the semantic front-end to a deterministic rule
engine~\cite{liu2024llava,dettmers2023qlora}, combining interpretability
with the expressive power of neural language understanding.

\subsection{Automation and Robotics in the Construction Domain}

The International Symposium on Automation and Robotics in Construction
(ISARC) provides a focused forum for construction-domain AI, and
contributions from ISARC~2023--2025 reveal three convergent themes
directly relevant to this paper: (i)~language-model-driven extraction
from inspection documents, (ii)~multimodal report generation from visual
data, and (iii)~deep-learning-based structural damage detection coupled
with maintenance scheduling.

\textbf{Language models for inspection document processing.}
Omar and Moselhi~\cite{omar2023bridge} (ISARC~2023) established an
integrated pipeline for automated acquisition and parsing of bridge
inspection reports, demonstrating that text-mining can replace manual
data entry for deck-deficiency records.
Their follow-on study~\cite{omar2025bridge} (ISARC~2025) replaced
rule-based parsing with fine-tuned Generative Pre-trained Transformers,
achieving substantial accuracy gains in recognising concrete-defect
entities from free-text engineering records --- a direct antecedent of
the QLoRA-based Damage VLM strategy presented here, which adapts the
same fine-tuning paradigm to Japanese inspection vocabulary.
Reja et al.~\cite{reja2025llm} (ISARC~2025) compared few-shot LLM
classifiers against fine-tuned transformers for road maintenance log
classification, confirming that domain-adapted language models
consistently outperform general-purpose prompting in specialised
infrastructure contexts.

\textbf{Multimodal and vision-language approaches.}
Pu et al.~\cite{pu2024mllm} (ISARC~2024) employed multimodal large
language models with Set-of-Mark prompting to generate construction
inspection reports directly from annotated site photographs --- the
closest prior work to our Damage VLM pipeline, though targeting
general construction quality in English rather than bridge-specific
damage in Japanese regulatory vocabulary.
Hsu et al.~\cite{hsu2024vlcon} (ISARC~2024) released VL-Con, a
construction-domain vision-language dataset, reinforcing the finding
that task-specific data collection is a prerequisite for reliable VLM
performance in AEC applications and motivating the curated Japanese
instruction dataset described in Section~\ref{sec:dataset}.
Mengiste et al.~\cite{mengiste2025progress} (ISARC~2025) automated
weekly construction progress reporting using multimodal LLM workflows,
further illustrating the industrial demand for language-based interfaces
to visual inspection data.

\textbf{Structural damage detection and maintenance decision support.}
Assad et al.~\cite{assad2025crack} (ISARC~2025) benchmarked
single-stage versus two-stage detectors for bridge crack detection
and segmentation, confirming that pixel-level localisation methods
remain the baseline for damage mapping, though without generating the
natural-language descriptions required for regulatory reporting.
Zuo et al.~\cite{zuo2025transformer} (ISARC~2025) combined
transformer-based detection with multi-resolution 3D reconstruction
to spatially localise bridge surface defects, demonstrating the
complementary role of geometric modelling alongside 2D image-based
language generation.
N\'{u}\~{n}ez Varillas et al.~\cite{nunez2024bridge} (ISARC~2024)
integrated UAV photogrammetry with CNN-based concrete damage
classification on reconstructed bridge 3D models, illustrating
how remote sensing reduces inspector exposure while maintaining
detection accuracy.
At the maintenance-decision level, Alsharqawi et al.~\cite{alsharqawi2025mrr}
(ISARC~2025) proposed an automated scheduling tool for bridge
maintenance, repair, and replacement using multi-objective genetic
algorithms.
Their portfolio-level optimiser is complementary to our inspection-level
priority scoring engine: the five-level repair priority index produced
by the ScoreBot provides the per-element damage state that such
optimisers require as input.

Collectively, these ISARC contributions confirm two transitions in
construction inspection practice: automated language processing is
displacing manual report writing, and multimodal architectures are
superseding single-modality vision pipelines.
The present work advances both transitions by fine-tuning a
state-of-the-art open-source VLM (LLaVA-1.5-7B) on a curated Japanese
bridge inspection dataset, delivering domain-appropriate structured
damage descriptions without dependence on proprietary APIs or
manual post-processing.

\subsection{AI Agent and Quality Guardrail}
\label{sec:related:guard}

The reliability of VLM-generated descriptions in safety-critical domains
critically depends on robust output quality control.
Three complementary lines of research are directly relevant to the
Quality Guard introduced in this work.

\textbf{LLM-as-Judge and quality evaluation.}
Zheng et al.~\cite{zheng2023judge} introduced the LLM-as-a-Judge paradigm
through MT-Bench and Chatbot Arena, demonstrating that a strong language
model can serve as an effective automated quality evaluator for free-text
outputs.
Building on this, Liu et al.~\cite{liu2023geval} proposed G-Eval, a
framework that employs GPT-4 with chain-of-thought to evaluate natural
language generation quality with higher human alignment than conventional
n-gram metrics.
Min et al.~\cite{min2023factscore} introduced FActScore to measure the
factual precision of long-form text by decomposing outputs into atomic
facts and verifying each independently.
These judge-based evaluation frameworks motivate using a fine-tuned SLM
(Swallow-8B~\cite{tokyotech2024swallow}) as a domain-aware quality
evaluator for VLM damage descriptions, following the principle of
``LLM judging LLM'' without relying on proprietary external APIs.

\textbf{Guardrails and self-correction in LLM pipelines.}
Constitutional AI~\cite{bai2022constitutional} proposed training AI
systems to critique and revise their own outputs according to a set of
principles, establishing the self-correction paradigm that underlies
quality guardrail architectures.
NeMo Guardrails~\cite{rebedea2023nemo} formalised this as a programmable
rail system, allowing developers to define topical, safety, and format
constraints that intercept LLM outputs before they reach downstream
applications --- directly analogous to the quality gate in our v0.6 pipeline.
Self-RAG~\cite{asai2023selfrag} extended the concept to retrieval-augmented
generation, where a model learns to reflect on whether retrieval and
generation steps should be invoked, achieving selective output through
special reflection tokens.
Chain-of-Verification~\cite{dhuliawala2023chainveri} reduced hallucination
in LLM outputs by generating independent verification questions and
cross-checking initial responses against them.
These approaches collectively inform our two-stage guard design:
Stage~1 applies deterministic rules (token-count thresholds and
repetition detection) analogous to structural rails;
Stage~2 invokes Swallow-8B as an LLM judge for borderline cases,
combining speed with semantic depth.

\textbf{AI agents with self-reflection.}
ReAct~\cite{yao2023react} combined reasoning traces with action steps
in language agents, enabling dynamic adjustment of the inference
trajectory based on observed feedback.
Reflexion~\cite{shinn2023reflexion} built on this by introducing verbal
reinforcement signals from a linguistic reflection mechanism, allowing
agents to revise strategies across trials without gradient updates.
Shen et al.~\cite{shen2023alignment} surveyed alignment techniques for
large language models, identifying output quality guardrails as a key
mechanism for deploying LLMs in high-stakes applications.

In the bridge inspection domain, the consequence of false-high quality
scores is misallocated maintenance budgets and deferred safety-critical
repairs.
By interposing an SLM judge between the VLM and the rule-based priority
scoring engine, the Quality Guard addresses the domain-specific risk that
VLMs trained on limited data may produce plausible but content-sparse
descriptions for images that are overexposed, occluded, or
low-resolution --- cases where assigning a priority score would be
misleading rather than informative.
Our guard distinguishes two failure modes: \textit{``Not recognised from
only image''} (insufficient visual information for any damage
description) and \textit{``Dirty or Noisy image''} (repetitive or
incoherent output). Only descriptions passing both stages receive a
five-level repair priority score; failed samples return
\textit{``No Score due to Low Quality Image''}, preserving the
integrity of the priority ranking for downstream maintenance planning.

%% ====================================================================
\section{Methodology}
\label{sec:methodology}
%% ====================================================================

\subsection{System Architecture Overview}

\begin{figure}[t]
\centering
\begin{tikzpicture}[
  node distance=0.65cm and 0.1cm,
  box/.style={rectangle, draw=gray!70, rounded corners=3pt,
              minimum width=3.2cm, minimum height=0.65cm,
              align=center, font=\footnotesize},
  stage/.style={box, fill=blue!12, draw=blue!50},
  proc/.style={box,  fill=orange!12, draw=orange!50},
  score/.style={box, fill=red!10,  draw=red!40},
  outbox/.style={box,   fill=green!12, draw=green!50},
  arr/.style={->, >=stealth, thick, gray!70},
  label/.style={font=\scriptsize\itshape, text=gray!70}
]

\node[stage]  (inp)  {\textbf{Stage 1}\\Image Input};
\node[proc,   below=of inp]  (vlm)  {\textbf{Stage 2}\\VLM Inference};
\node[proc,   below=of vlm]  (ext)  {\textbf{Stage 3}\\Structured Extraction};
\node[score,  below=of ext]  (scr)  {\textbf{Stage 4}\\Priority Scoring};
\node[outbox,    below=of scr]  (outnd)  {Repair Priority\\1--5 Index};

\draw[arr] (inp) -- (vlm);
\draw[arr] (vlm) -- (ext);
\draw[arr] (ext) -- (scr);
\draw[arr] (scr) -- (outnd);

\node[label, right=0.15cm of inp] {800 test images};
\node[label, right=0.15cm of vlm] {LLaVA-1.5-7B QLoRA};
\node[label, right=0.15cm of ext] {JSON: member, damage};
\node[label, right=0.15cm of scr] {Severity/Type/Location};
\end{tikzpicture}
\caption{End-to-end pipeline for bridge damage understanding and repair
         priority scoring. Fine-tuning (Section~\ref{sec:finetuning}) targets
         Stage~2; Stages~3--4 remain rule-based.}
\label{fig:pipeline}
\end{figure}
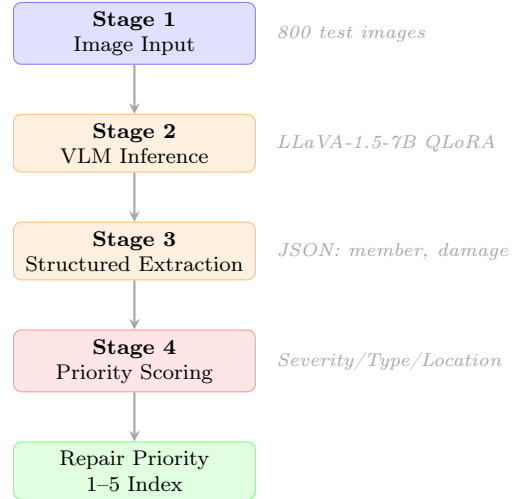

Figure~\ref{fig:pipeline} illustrates the end-to-end pipeline comprising four
sequential stages: image input, VLM-based damage understanding, structured
information extraction, and repair priority scoring.
A detailed algorithm flow that incorporates the Quality Guard Agent
is shown in Figure~\ref{fig:algorithm_flow} (Section~\ref{sec:quality_guard}).

%% ------ TikZ Pipeline Figure ------

%% ====================================================================
\subsection{Dataset Construction}
\label{sec:dataset}
%% ====================================================================

\subsubsection{Source Data}

We collect 10{,}789 paired bridge damage images and Japanese inspection text
records from routine close-up visual inspections conducted under Japan's Road
Act, which mandates five-year periodic bridge inspections.
Each record contains a damage image and the corresponding inspector's \emph{shooken}
(observation text, Japanese: \textit{shoken}) written using standardised civil engineering terminology.

\subsubsection{Quality Filtering}

To ensure training data quality, we apply the following filters:
\begin{itemize}[leftmargin=*,itemsep=1pt]
  \item \textbf{Text length}: 15--500 characters (removes empty and excessively
        verbose records).
  \item \textbf{Invalid patterns}: excludes records containing only placeholder
        text (``nashi'' (none), ``???'', or similar).
  \item \textbf{Keyword requirement}: at least one structural member term
        (e.g., \textit{girder, deck, pier}) and one damage term
        (e.g., \textit{rebar exposure, crack, corrosion}) must be present.
\end{itemize}

\subsubsection{Stratified Sampling and Split Strategy}

After filtering, we construct training and evaluation sets using stratified
sampling by cross-product of member type and damage type to ensure class balance.

\begin{itemize}[leftmargin=*,itemsep=1pt]
  \item \textbf{Fixed test set}: 800 samples drawn with random seed 42, held
        out for all experiments (798 with valid images, 2 missing).
  \item \textbf{Progressive training sets}: independently sampled sets of
        1k, 2k, 3k, and 4k images to study data-scaling behaviour.
        Each training set uses an 80/20 train/validation split.
        This progression follows the scale-search methodology
        in~\cite{yasuno2026japanese}, which identified $n\approx4{,}000$
        as optimal for Japanese domain-specific language models---extended
        here to the multimodal VLM setting.
\end{itemize}

Table~\ref{tab:dataset} shows the resulting dataset statistics.

\begin{table}[h]
\centering
\caption{Dataset statistics after quality filtering.}
\label{tab:dataset}
\begin{tabular}{lrr}
\toprule
\textbf{Split} & \textbf{Images} & \textbf{Valid (\%)} \\
\midrule
Test (fixed)          & 800   & 798 (99.75\%) \\
\midrule
Train-1k (train/val)  & 799 / 200   & 100\% \\
Train-2k (train/val)  & 1{,}599 / 400 & 100\% \\
Train-3k (train/val)  & 2{,}398 / 600 & 100\% \\
Train-4k (train/val)  & 3{,}196 / 799 & 100\% \\
\midrule
Source pool           & 10{,}789 & --- \\
\bottomrule
\end{tabular}
\end{table}

%% ====================================================================
\subsection{QLoRA Fine-Tuning}
\label{sec:finetuning}
%% ====================================================================

\subsubsection{Base Model}

We fine-tune \textbf{LLaVA-1.5-7B}~\cite{liu2024llava} (HuggingFace identifier:
\texttt{llava-hf/llava-1.5-7b-hf}), an instruction-tuned vision-language model
combining CLIP ViT-L/14 as the vision encoder and Vicuna-1.5-7B as the
language decoder connected via a two-layer MLP projection.
LLaVA-1.5-7B is chosen for its strong visual instruction-following capability
and practical size for consumer GPU deployment.

\subsubsection{QLoRA Configuration}

We apply \textbf{QLoRA}~\cite{dettmers2023qlora} (Quantized Low-Rank Adaptation),
which loads the base model in 4-bit NF4 precision while training low-rank adapter
matrices in full precision. This enables fine-tuning a 7B-parameter model within
16GB VRAM.

The quantization configuration uses 4-bit NF4 with double quantization (compressing
quantization constants themselves) via \texttt{bitsandbytes}~\cite{dettmers2022llmint8}.
Low-rank adapters are inserted into all major projection matrices as shown in
Table~\ref{tab:qlora_config}.

\begin{table}[h]
\centering
\caption{QLoRA and training hyperparameters.}
\label{tab:qlora_config}
\begin{tabular}{ll}
\toprule
\textbf{Parameter} & \textbf{Value} \\
\midrule
\multicolumn{2}{l}{\textit{Quantization}} \\
Precision & 4-bit NF4 \\
Double quantization & Enabled \\
Compute dtype & float16 \\
\midrule
\multicolumn{2}{l}{\textit{LoRA Adapters}} \\
Rank $r$ & 32 \\
Alpha $\alpha$ & 64 \\
Dropout & 0.05 \\
Target modules & \texttt{q,k,v,o\_proj;} \\
 & \texttt{gate,up,down\_proj} \\
\midrule
\multicolumn{2}{l}{\textit{Training}} \\
Batch size & 4 \\
Gradient accumulation & 4 (eff.\ batch = 16) \\
Learning rate & $2 \times 10^{-4}$ \\
Epochs & 3 \\
Warmup steps & 50 \\
Max gradient norm & 1.0 \\
Weight decay & 0.01 \\
Optimizer & AdamW \\
LR scheduler & Cosine \\
Mixed precision & FP16 \\
\midrule
\multicolumn{2}{l}{\textit{Hardware}} \\
GPU & RTX 4060 Ti 16GB \\
CUDA & 12.4 \\
PyTorch & 2.6.0+cu124 \\
Transformers & 4.57.6 \\
PEFT & 0.18.1 \\
\bottomrule
\end{tabular}
\end{table}

The adapter scaling factor is $s = \alpha / r = 64/32 = 2$, controlling the
magnitude of the learned residual updates relative to the frozen base weights.
The LoRA decomposition for each weight matrix $W_0 \in \mathbb{R}^{d \times k}$ is:
\begin{equation}
  W = W_0 + \frac{\alpha}{r} \, B A
  \label{eq:lora}
\end{equation}
where $B \in \mathbb{R}^{d \times r}$ and $A \in \mathbb{R}^{r \times k}$ are the
low-rank adapter matrices initialised as $B = \mathbf{0}$ and $A \sim \mathcal{N}(0, \sigma^2)$.

\subsubsection{Progressive Training Strategy}
\label{sec:progressive}

To investigate data-scaling behaviour, we train four independent models on
progressively larger datasets (1k, 2k, 3k, 4k) using identical hyperparameters.
Table~\ref{tab:training_results} reports validation loss and training duration
for each scale.

\begin{table}[h]
\centering
\caption{Progressive training results (LLaVA-1.5-7B QLoRA).}
\label{tab:training_results}
\begin{tabular}{lrrl}
\toprule
\textbf{Data}& \textbf{Val Loss} & \textbf{Duration} & \textbf{Best Ckpt} \\
\midrule
1k & 3.135 & 1:22:57 & step-100/150 \\
2k & 3.073 & 2:55:37 & step-300/300 \\
3k & 3.073 & 4:31:44 & step-400/450 \\
4k & 3.067 & 6:19:10 & step-600/600 \\
\bottomrule
\end{tabular}
\end{table}

Three key observations emerge:
\begin{enumerate}[leftmargin=*,itemsep=1pt]
  \item \textbf{1k $\to$ 2k}: Validation loss decreases by 1.98\%,
        the largest single-step improvement, confirming benefit of additional data.
  \item \textbf{Plateau at 2k--3k}: Loss stagnates at 3.073 despite 1.6$\times$
        more training time, indicating convergence of the fine-tuning signal.
  \item \textbf{Marginal gain at 4k}: Only 0.2\% improvement over 3k at
        1.4$\times$ cost, demonstrating strong diminishing returns.
\end{enumerate}

The \textbf{2k model} therefore offers the optimal cost--benefit trade-off:
near-maximum accuracy ($\Delta = 0.2\%$ from 4k) achieved in 2.9 hours---2.2$\times$
faster than the 4k baseline.

%% ====================================================================
\subsection{Inference Optimisation}
\label{sec:inference}
%% ====================================================================

Running inference on 800 test images with an unoptimised HuggingFace
\texttt{generate()} call requires 33.79 seconds per image (7.5 hours total),
making systematic multi-model evaluation impractical. We implement three
complementary optimisations.

\subsubsection{\texttt{torch.compile()} Graph Optimisation}

PyTorch 2.0 introduced \texttt{torch.compile()}~\cite{pytorch2023compile},
which applies TorchDynamo bytecode analysis and TorchInductor ahead-of-time
compilation to the model's computation graph.
We use \texttt{mode=\textrm{``}reduce-overhead\textrm{''}}, which
minimises Python interpreter overhead---the dominant bottleneck for repeated
single-batch inference calls.
An initial warm-up batch ($\approx$60--80 seconds) is followed by accelerated
subsequent inference.

\subsubsection{Batch Processing}

We implement a batched inference loop that processes $B$ images simultaneously,
padding inputs to uniform length within each batch:

\begin{equation}
  \hat{T}_{\text{total}} = \underbrace{t_{\text{compile}}}_{\text{once}} +
  \left\lceil \frac{N}{B} \right\rceil \cdot t_{\text{batch}}(B)
  \label{eq:batch_time}
\end{equation}
where $N = 800$ is the test set size and $t_{\text{batch}}(B)$ is the
per-batch latency at batch size $B$.

\subsubsection{Batch Size Selection}

We empirically evaluate batch sizes $B \in \{4, 8, 16\}$ on the RTX 4060 Ti
(16GB VRAM). Table~\ref{tab:batch_opt} reports per-image latency and GPU memory
utilisation.

\begin{table}[h]
\centering
\caption{Batch size optimisation results (2k model, 16 test samples).}
\label{tab:batch_opt}
\begin{tabular}{lrrl}
\toprule
\textbf{Batch} & \textbf{s/image} & \textbf{VRAM} & \textbf{Status} \\
\midrule
Baseline (v0.4) & 33.79 & --- & Reference \\
$B=4$ & 16.80 & $<$80\% & \checkmark \\
$B=8$ & 10.06 & $\approx$78\% & \checkmark\; \textbf{Optimal} \\
$B=16$ & --- & 98\% & OOM risk \\
\bottomrule
\end{tabular}
\end{table}

$B = 8$ reduces per-image latency by 70.2\% relative to the baseline (33.79 $\to$
10.06 seconds) while keeping GPU memory below the 80\% safety threshold.
$B = 16$ saturates VRAM at 15.7/16.0 GB, inducing long \texttt{torch.compile()}
compilation and out-of-memory risk; it is therefore excluded.

\subsubsection{Token Budget Optimisation}

We analyse per-model output token distributions on 40-sample pilot runs to set
\texttt{max\_new\_tokens} without systematic truncation:

\begin{table}[h]
\centering
\caption{Output token statistics by training scale.}
\label{tab:tokens}
\begin{tabular}{lrrc}
\toprule
\textbf{Model} & \textbf{Mean} & \textbf{Std} & \textbf{max new tokens} \\
\midrule
1k & 348.5 & 148.4 & 512 \\
2k & 279.7 &  84.0 & 384 \\
3k & 266.4 & 160.0 & 384 \\
4k & 231.6 &  70.8 & 384 \\
\bottomrule
\end{tabular}
\end{table}

More training data produces more concise outputs (33.5\% token reduction from
1k to 4k), an inverse-scaling phenomenon we attribute to improved instruction
following. Setting \texttt{max\_new\_tokens}$=384$ eliminates truncation for 2k
and 4k models (10\% and 0\% truncation, respectively) while reducing generation
time by 25\% over the conservative default of 512.

We additionally apply \texttt{repetition\_penalty}$=1.2$ to suppress
the looping behaviour observed in 1k model outputs.

\subsubsection{Final Optimised Configuration}

The combined optimisation (v0.5.1) achieves:
\begin{itemize}[leftmargin=*,itemsep=1pt]
  \item \textbf{10.06 s/image} (batch\_size=8, compiled)
  \item \textbf{2.2 hours} for 800 images per model
  \item \textbf{6.7 hours} for three models (2k/3k/4k) combined
\end{itemize}

compared to 22.5 hours with the unoptimised v0.4 script---a 70.2\% wall-clock
reduction enabling overnight evaluation of all four model scales.

%% ====================================================================
\subsection{Structured Information Extraction}
\label{sec:extraction}
%% ====================================================================

The VLM outputs free-form Japanese text describing the damage scene.
We extract five structured attributes via a rule-based parser:

\begin{equation}
  j_i = \bigl(\underbrace{m_i}_{\text{member}},\;
               \underbrace{d_i}_{\text{damage}},\;
               \underbrace{l_i}_{\text{location}},\;
               \underbrace{v_i}_{\text{severity}},\;
               \underbrace{e_i}_{\text{extent}}\bigr)
  \label{eq:struct}
\end{equation}

Extraction uses a two-pass strategy: (1) keyword matching against a
civil-engineering vocabulary dictionary to identify candidate spans, and
(2) negation-aware pattern filtering to exclude ``no damage'' clauses.
Table~\ref{tab:extract_fields} defines each attribute and its vocabulary.

\begin{table}[h]
\centering
\caption{Structured extraction attributes.}
\label{tab:extract_fields}
\setlength{\tabcolsep}{3pt}
\begin{tabular}{lp{4.0cm}}
\toprule
\textbf{Attribute} & \textbf{Example values} \\
\midrule
Member ($m$)   & girder, deck, pier, bearing, railing \\
Damage ($d$)   & rebar exposure, crack, corrosion, spalling, section loss \\
Location ($l$) & bottom face, upper surface, joint, support \\
Severity ($v$) & high, medium, low \\
Extent ($e$)   & local, widespread, 30\%, 1.5\,m$^2$ \\
\bottomrule
\end{tabular}
\end{table}

%% ====================================================================
\subsection{Repair Priority Scoring System}
\label{sec:scoring}
%% ====================================================================

\subsubsection{Score Computation}

Given structured attributes $j_i$, the repair priority score
$s_i \in [0, 1]$ is computed as a weighted sum:

\begin{equation}
  s_i = w_d\,\phi_d(d_i) + w_v\,\phi_v(v_i) + w_l\,\phi_l(l_i) + w_r\,\phi_r(r_i)
  \label{eq:score}
\end{equation}

where $\phi_*: \text{category} \to [0,1]$ are look-up functions defined in
a YAML rule file, and the weights are:
\begin{equation}
  (w_d,\, w_v,\, w_l,\, w_r) = (0.35,\; 0.40,\; 0.15,\; 0.10)
  \label{eq:weights}
\end{equation}

with $w_d + w_v + w_l + w_r = 1.0$.
Severity carries the highest weight (40\%) reflecting its direct relationship
to structural safety; damage type is second (35\%) because certain damage modes
(rebar exposure, section loss) carry inherent structural risk regardless of
assessed severity.

\subsubsection{Damage-Type and Location Scores}

Table~\ref{tab:scoring_rules} lists key look-up values derived from
Japanese bridge inspection guidelines~\cite{mlit2023bridge}.

\begin{table}[h]
\centering
\caption{Selected scoring rule look-up values.}
\label{tab:scoring_rules}
\begin{tabular}{lrl}
\toprule
\textbf{Category} & $\phi$ & \textbf{Note} \\
\midrule
\multicolumn{3}{l}{\textit{Damage type $\phi_d$}} \\
Section loss       & 1.00 & Structural emergency \\
Rebar exposure     & 0.95 & Corrosion propagation \\
Corrosion          & 0.85 & Durability risk \\
Spalling           & 0.75 & Cover loss \\
Crack              & 0.60 & Context-dependent \\
Stain              & 0.30 & Aesthetic only \\
\midrule
\multicolumn{3}{l}{\textit{Location $\phi_l$}} \\
Girder bottom face & 1.00 & Primary load path \\
Girder             & 0.95 & \\
Bearing            & 0.90 & Load transfer \\
Expansion joint    & 0.80 & Water ingress \\
Deck               & 0.75 & \\
Railing            & 0.40 & Non-structural \\
\midrule
\multicolumn{3}{l}{\textit{Severity $\phi_v$}} \\
High               & 1.00 & \\
Medium             & 0.60 & \\
Low                & 0.30 & \\
\bottomrule
\end{tabular}
\end{table}

\subsubsection{Combination Bonuses}

Certain co-occurring attribute pairs trigger additive bonuses $\delta$ to
capture compound structural risk that weighted summation alone may underestimate:

\begin{equation}
  s_i' = \min\!\left(1.0,\; s_i + \sum_k \delta_k \cdot \mathbb{1}[\text{cond}_k(j_i)]\right)
  \label{eq:bonus}
\end{equation}

For example: girder-bottom section loss ($\delta=+0.10$), high-severity rebar
exposure ($\delta=+0.08$), and bearing corrosion ($\delta=+0.07$).

\subsubsection{Five-Level Priority Index}

The continuous score $s_i'$ is discretised into a five-level repair
priority index:

\begin{equation}
  P_i = \begin{cases}
    5 & s_i' \geq 0.85 \quad \text{(Immediate)} \\
    4 & s_i' \geq 0.70 \quad \text{(High)} \\
    3 & s_i' \geq 0.50 \quad \text{(Moderate)} \\
    2 & s_i' \geq 0.35 \quad \text{(Low)} \\
    1 & s_i' <  0.35 \quad \text{(Minimal)}
  \end{cases}
  \label{eq:priority}
\end{equation}

Priority 5 triggers immediate-repair alerts; Priority 1 records a finding
without scheduling intervention.

%% ====================================================================
\subsection{Quality Guard Agent}
\label{sec:quality_guard}
%% ====================================================================

To prevent low-quality VLM outputs from generating spurious repair priority
scores, we interpose a two-stage \textbf{Quality Guard Agent} between the
VLM inference and the scoring engine.
Only descriptions that pass both stages receive a five-level priority score;
failed samples return \textit{``No Score due to Low Quality Image''}, preserving
the integrity of the priority ranking for downstream maintenance planning.
Figure~\ref{fig:algorithm_flow} illustrates the complete algorithm flow
of the Visual Inspection ScoreBot~v0.6.3 pipeline.

\subsubsection{Empirical Threshold Derivation}

\begin{figure}[h]
\centering
\includegraphics[width=0.46\textwidth]{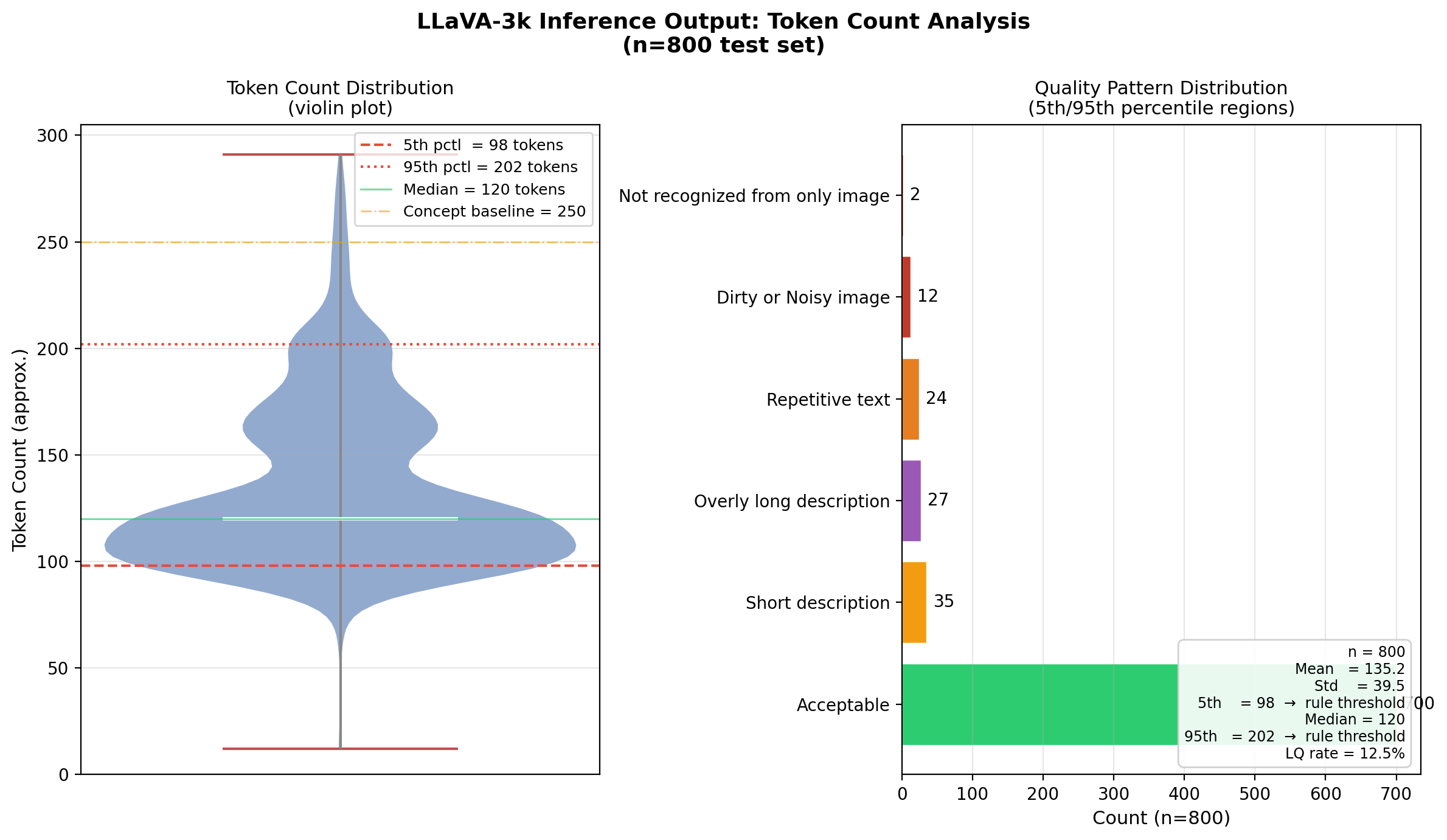}
\caption{LLaVA-3k inference output token analysis ($n=800$).
\textit{Left}: violin plot showing the token-count distribution
(mean\,=\,135.2, std\,=\,39.5, median\,=\,120).
Red dashed/dotted lines mark the empirically derived rule thresholds
($\theta_{\text{low}}=98$ at 5th pct; $\theta_{\text{high}}=202$ at 95th pct).
\textit{Right}: quality pattern breakdown within the threshold regions.
Acceptable outputs (87.5\%) dominate; low-quality patterns total 12.5\%
across five failure categories.}
\label{fig:lq_analysis}
\end{figure}

Rather than setting token-count thresholds from domain heuristics alone,
we first analyse the distribution of output lengths over all $n=800$
test-set predictions.
Figure~\ref{fig:lq_analysis} presents (left) a violin plot of the token-count
distribution and (right) the breakdown of low-quality patterns within the
5th and 95th percentile tails.

The violin shape reveals a principal density mass between 100 and 220~tokens
with a narrow waist below 100, indicating that genuinely informative damage
descriptions cluster in a well-defined range.
We set the token thresholds at empirical percentiles:

\begin{equation}
  \theta_{\text{low}} = 98 \;\text{(5th pct)}, \quad
  \theta_{\text{high}} = 202 \;\text{(95th pct)}
  \label{eq:thresholds}
\end{equation}

\noindent
This data-driven calibration avoids the twin risks of an over-tight lower bound
(rejecting valid short descriptions) and an over-loose upper bound
(admitting repetitive or hallucinated long outputs).
Table~\ref{tab:lq_patterns} summarises the quality-pattern breakdown.

\begin{table}[h]
\centering
\caption{Low-quality pattern breakdown ($n=800$, 3k model).}
\label{tab:lq_patterns}
\setlength{\tabcolsep}{4pt}
\begin{tabular}{lrr}
\toprule
\textbf{Pattern} & \textbf{Count} & \textbf{\%} \\
\midrule
Acceptable                     & 700 & 87.5 \\
Short description              &  35 &  4.4 \\
Overly long description        &  27 &  3.4 \\
Repetitive text                &  24 &  3.0 \\
Dirty or Noisy image           &  12 &  1.5 \\
Not recognised from only image &   2 &  0.2 \\
\midrule
\textbf{Low-quality total}     & 100 & 12.5 \\
\bottomrule
\end{tabular}
\end{table}

\subsubsection{Scope Definition}

A critical design decision is the evaluation scope.
As introduced in Section~\ref{sec:introduction}, this work targets
aspect~(1) of damage understanding---current-state observation from a
single image---so the Quality Guard is calibrated against this scope
rather than against multi-period or causal reasoning.
Inspection text generated by VLMs trained on real inspection records
tends to reproduce the reporting style of the training data, including
references to previous inspection comparisons
(\textit{``since the last inspection\ldots''}) and maintenance
recommendations (\textit{``corrective action is required''}).
These out-of-scope clauses are artefacts of the training distribution,
not hallucinations in the strict sense.

We therefore operationalise the scope as a \textit{``one image,
current time limited''} criterion:
quality is assessed exclusively on the current-state damage observation.
Out-of-scope content (two-time-point comparisons, maintenance directives)
is \textbf{not penalised} provided that current damage information is
present; only descriptions that consist \textit{entirely} of out-of-scope
content, with no damage information for the current image, are classified
as FAIL.
This design preserves useful damage signals while filtering outputs that
carry no actionable inspection content.
The broader implications of this scope choice, including extensions to
aspects~(2)--(4), are revisited in Section~\ref{sec:disc_panel}.

\subsubsection{Stage 1: Rule-Based Filter}
\label{sec:guard_stage1}

Stage~1 applies four deterministic rules to each prediction text $\hat{t}_i$
at negligible cost (${\approx}0.01$~s per sample, CPU-only):

\begin{enumerate}[leftmargin=*,itemsep=1pt]
  \item \textbf{File existence} (Stage~0 pre-filter): if the source image is
        absent, return \textit{``No such file or directory''} (FAIL).
  \item \textbf{Token count}: if $|\hat{t}_i| < \theta_{\text{low}} = 98$,
        return \textit{``Short description''} (FAIL).
  \item \textbf{Repetition detection}: if a repeated $n$-gram pattern exceeds
        a coverage threshold, return \textit{``Dirty or Noisy image''} (FAIL).
  \item \textbf{Keyword absence}: if no structural damage keyword is present
        and vague, out-of-scope phrases dominate, return
        \textit{``Not recognised from only image''} (FAIL).
\end{enumerate}

Predictions passing all four rules proceed to Stage~2.
Stage~1 alone captures 90.9\% of outputs as \textit{High Quality (rule only)}
at a throughput of ${\approx}0.01$~s/row, enabling rapid triage without GPU
resources.

\subsubsection{Stage 2: SLM-as-Judge}
\label{sec:guard_stage2}

Stage~2 invokes \textbf{Swallow-8B}~\cite{tokyotech2024swallow}
(\path{tokyotech-llm/Llama-3-Swallow-8B-Instruct-v0.1}),
a Japanese instruction-tuned SLM, as a semantic quality judge following
the LLM-as-a-Judge paradigm~\cite{zheng2023judge}.
The model receives the extracted text with a structured prompt specifying
the evaluation scope and returns a verdict in a fixed two-line format:

\begin{small}
\begin{verbatim}
VERDICT: <PASS | FAIL>
REASON_CODE: <High Quality | Short description
             | Dirty or Noisy image
             | Not recognised from only image>
\end{verbatim}
\end{small}

Inference runs under the \textbf{Unsloth}~\cite{han2023unsloth} framework
(\texttt{FastLanguageModel.from\_pretrained} + \texttt{for\_inference()})
with 4-bit NF4 quantisation, fitting within the 16\,GB VRAM budget after
LLaVA-1.5-7B is unloaded.
Batch inference at $B=8$ (Section~\ref{sec:inference}) achieves
\textbf{10.10~seconds per guarded image} end-to-end on the RTX\,4060\,Ti.

The system prompt explicitly states the one-image/current-time scope
and instructs the judge to disregard out-of-scope clauses when
evaluating whether damage content is present.

\subsubsection{Quality Guard Outcomes}

Table~\ref{tab:qg_outcomes} reports the final Quality Guard Agent
outcomes over all $n=800$ test samples from the v0.6.3 pipeline.
All failures are captured at Stage~1 (rule-based filter); Stage~2
(Swallow-8B SLM judge) confirms all 727 Stage~1 PASS samples as
\textit{High Quality}, adding no additional rejections.
This indicates that the empirically calibrated rule thresholds
(Section~\ref{sec:guard_stage1}) already cover the observable
low-quality patterns in the 3k model's output distribution.

\begin{table}[h]
\centering
\caption{Quality Guard Agent final outcomes ($n=800$, 3k model,
         v0.6.3 pipeline). All FAILs originate at Stage~1;
         Stage~2 yields no additional rejections.}
\label{tab:qg_outcomes}
\setlength{\tabcolsep}{3pt}
\begin{tabular}{lrr}
\toprule
\textbf{Outcome} & \textbf{Count} & \textbf{\%} \\
\midrule
PASS --- High Quality          & 727 & 90.9 \\
FAIL --- Dirty or Noisy image  &  36 &  4.5 \\
FAIL --- Short description     &  35 &  4.4 \\
FAIL --- No such file          &   2 &  0.2 \\
\midrule
\textbf{FAIL total}            &  73 &  9.1 \\
\bottomrule
\end{tabular}
\end{table}

%% ====================================================================
\subsection{Evaluation Protocol}
\label{sec:evaluation}
%% ====================================================================

\subsubsection{Vector Similarity Metric}

Rather than exact string matching, we evaluate generation quality using
\textbf{cosine similarity in semantic embedding space}.
Ground-truth inspection texts and model predictions are encoded with
\path{sonoisa/sentence-bert-base-ja-mean-tokens-v2}~\cite{sonoisa2021sbert},
a Japanese Sentence-BERT model producing 768-dimensional sentence vectors
pre-trained on Japanese Wikipedia and SNLI-translated corpora.

\begin{equation}
  \text{sim}(t_i^{\text{gt}}, t_i^{\text{pred}}) =
  \frac{\mathbf{e}_i^{\text{gt}} \cdot \mathbf{e}_i^{\text{pred}}}
       {\|\mathbf{e}_i^{\text{gt}}\| \;\|\mathbf{e}_i^{\text{pred}}\|}
  \label{eq:cosine}
\end{equation}

where $\mathbf{e} = \text{SentBERT}(t) \in \mathbb{R}^{768}$.
We report mean cosine similarity $\bar{\rho}$ and its standard deviation
$\sigma_\rho$ over all 798 valid test pairs.

\subsubsection{Quality Tier Classification}

We map continuous similarity scores to five quality tiers to provide
actionable assessments for model selection:

\begin{table}[h]
\centering
\caption{Quality tier classification thresholds.}
\label{tab:quality_tiers}
\begin{tabular}{lll}
\toprule
\textbf{Tier} & \textbf{Threshold} & \textbf{Interpretation} \\
\midrule
Excellent  & $\rho \geq 0.85$ & Near-expert match \\
Good       & $\rho \geq 0.75$ & Minor omissions only \\
Acceptable & $\rho \geq 0.65$ & Usable for triage \\
Poor       & $\rho \geq 0.50$ & Significant gaps \\
Very Poor  & $\rho <  0.50$   & Unreliable \\
\bottomrule
\end{tabular}
\end{table}

A model achieving $\bar{\rho} \geq 0.65$ (Acceptable tier) is considered
viable for production deployment as an AI-assisted triage tool.

\subsubsection{Baseline Reference}

The 1k model evaluated on the fixed 800-sample test set establishes the
baseline: $\bar{\rho}_{1k} = 0.6491 \pm 0.0875$, placing it in the
\textit{Acceptable} tier.
The 2k model achieves $\bar{\rho}_{2k} = 0.6850 \pm 0.0814$
(+5.5\% relative improvement), remaining in the \textit{Acceptable} tier
but approaching the \textit{Good} boundary ($\geq 0.70$).
Comparative results for the 3k and 4k models will be added upon
completion of inference.

\begin{table}[h]
\centering
\caption{Cosine similarity by training scale (800 test samples).
         Tier thresholds: Excellent$\geq$0.85, Good$\geq$0.70,
         Acceptable$\geq$0.60, Poor$\geq$0.50.}
\label{tab:results}
\setlength{\tabcolsep}{5pt}
\begin{tabular}{lrrlr}
\toprule
Model & $\bar{\rho}$ & $\sigma$ & Tier & s/image \\
\midrule
1k & 0.6491 & 0.0875 & Acceptable & 33.79 \\
2k & 0.6850 & 0.0814 & Acceptable & 10.06 \\
3k & 0.6909 & 0.0784 & Acceptable & 7.50 \\
4k & 0.6739 & 0.0795 & Acceptable & 6.17 \\
\bottomrule
\end{tabular}
\end{table}

%% ====================================================================
\subsection{Implementation and Reproducibility}
\label{sec:implementation}
%% ====================================================================

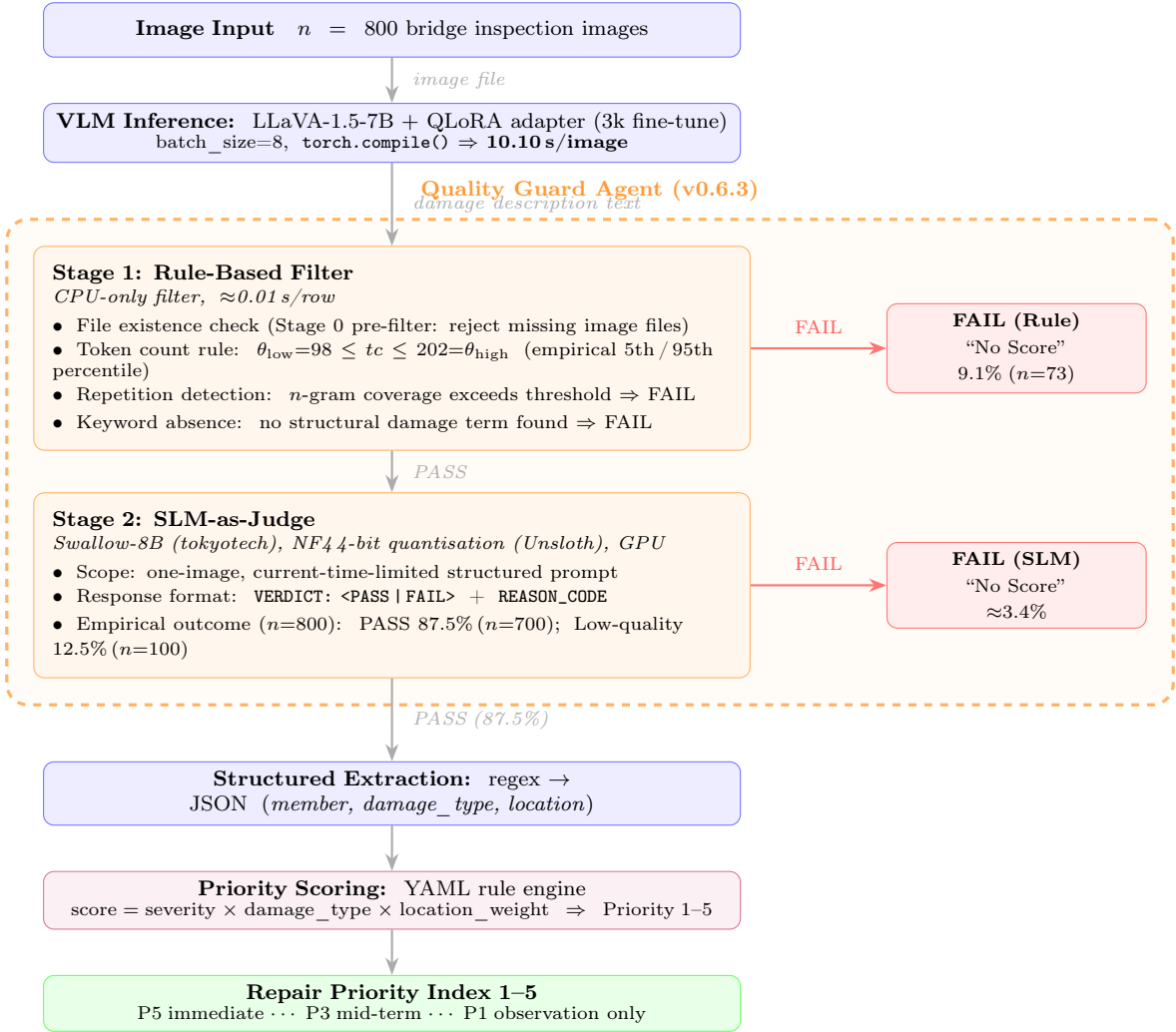
\begin{figure*}[t]
\centering
\begin{tikzpicture}[
  node distance=0.60cm,
  %% Main pipeline nodes (blue)
  pipe/.style={rectangle, rounded corners=4pt,
               draw=blue!55, fill=blue!7,
               text width=9.0cm, minimum height=0.72cm,
               align=center, font=\footnotesize},
  %% Quality Guard stage nodes (orange, left-aligned text)
  grd/.style={rectangle, rounded corners=4pt,
              draw=orange!60, fill=orange!7,
              text width=9.0cm,
              align=left, inner sep=7pt, font=\footnotesize},
  %% Scoring node (purple)
  scr/.style={rectangle, rounded corners=4pt,
              draw=purple!55, fill=purple!6,
              text width=9.0cm, minimum height=0.72cm,
              align=center, font=\footnotesize},
  %% Output node (green)
  otn/.style={rectangle, rounded corners=4pt,
              draw=green!60, fill=green!9,
              text width=9.0cm, minimum height=0.72cm,
              align=center, font=\footnotesize},
  %% Fail exit nodes (red, right column)
  fal/.style={rectangle, rounded corners=4pt,
              draw=red!55, fill=red!7,
              text width=3.2cm, minimum height=0.72cm,
              align=center, font=\scriptsize},
  %% Arrows
  arr/.style={-Stealth, thick, gray!65},
  far/.style={-Stealth, thick, red!55},
  %% Annotation labels
  alb/.style={font=\scriptsize\itshape, text=gray!65},
]

%% -------- Stage 1: Image Input --------
\node[pipe] (img)
  {\textbf{Image Input}\quad $n=800$ bridge inspection images};

%% -------- Stage 2: VLM Inference --------
\node[pipe, below=of img] (vlm)
  {\textbf{VLM Inference:}\; LLaVA-1.5-7B $+$ QLoRA adapter (3k fine-tune)\\[-2pt]
   {\scriptsize batch\_size=8,\; \texttt{torch.compile()} $\Rightarrow$ \textbf{10.10\,s/image}}};

%% -------- Quality Guard Stage 1 --------
\node[grd, below=1.1cm of vlm] (g1)
  {\textbf{Stage~1: Rule-Based Filter}\\[-1pt]
   {\scriptsize\itshape CPU-only filter,\; $\approx$0.01\,s/row}\\[3pt]
   {\scriptsize
    \textbullet\; File existence check (Stage~0 pre-filter: reject missing image files)\\[1pt]
    \textbullet\; Token count rule:\; $\theta_{\text{low}}{=}98 \leq tc \leq
                  202{=}\theta_{\text{high}}$\; (empirical 5th\,/\,95th percentile)\\[1pt]
    \textbullet\; Repetition detection:\; $n$-gram coverage exceeds threshold
                  $\Rightarrow$ FAIL\\[1pt]
    \textbullet\; Keyword absence:\; no structural damage term found $\Rightarrow$ FAIL}};

%% -------- Quality Guard Stage 2 --------
\node[grd, below=0.55cm of g1] (g2)
  {\textbf{Stage~2: SLM-as-Judge}\\[-1pt]
   {\scriptsize\itshape Swallow-8B (tokyotech), NF4\,4-bit quantisation (Unsloth), GPU}\\[3pt]
   {\scriptsize
    \textbullet\; Scope: one-image, current-time-limited structured prompt\\[1pt]
    \textbullet\; Response format:\;
                  \texttt{VERDICT: <PASS\,|\,FAIL>}\; +\; \texttt{REASON\_CODE}\\[1pt]
    \textbullet\; Empirical outcome ($n{=}800$):\;
                  PASS 87.5\%\,($n{=}700$);\; Low-quality 12.5\%\,($n{=}100$)}};

%% -------- Stage 3: Structured Extraction --------
\node[pipe, below=1.1cm of g2] (ext)
  {\textbf{Structured Extraction:}\;
   regex $\to$ JSON\; (\textit{member, damage\_type, location})};

%% -------- Stage 4: Priority Scoring --------
\node[scr, below=of ext] (scoring)
  {\textbf{Priority Scoring:}\; YAML rule engine\\[-2pt]
   {\scriptsize
    score\;$=$\;severity\;$\times$\;damage\_type\;$\times$\;location\_weight
    $\;\Rightarrow\;$ Priority\;1--5}};

%% -------- Output --------
\node[otn, below=of scoring] (out)
  {\textbf{Repair Priority Index 1--5}\\[-2pt]
   {\scriptsize
    P5\;immediate\;$\cdots$\;P3\;mid-term\;$\cdots$\;P1\;observation only}};

%% -------- FAIL exit nodes (right column) --------
\node[fal, right=1.8cm of g1] (f1)
  {\textbf{FAIL (Rule)}\\[2pt]``No Score''\\[2pt]{\scriptsize 9.1\%\;($n{=}73$)}};
\node[fal, right=1.8cm of g2] (f2)
  {\textbf{FAIL (SLM)}\\[2pt]``No Score''\\[2pt]{\scriptsize $\approx$3.4\%}};

%% -------- Main flow arrows --------
\draw[arr] (img)     -- (vlm)
  node[midway, right=4pt, alb] {image file};
\draw[arr] (vlm)     -- (g1)
  node[midway, right=4pt, alb] {damage description text};
\draw[arr] (g1)      -- (g2)
  node[midway, right=4pt, alb] {PASS};
\draw[arr] (g2)      -- (ext)
  node[midway, right=4pt, alb] {PASS\;(87.5\%)};
\draw[arr] (ext)     -- (scoring);
\draw[arr] (scoring) -- (out);

%% -------- FAIL branch arrows --------
\draw[far] (g1.east) -- (f1.west)
  node[midway, above=2pt, font=\scriptsize, text=red!65] {FAIL};
\draw[far] (g2.east) -- (f2.west)
  node[midway, above=2pt, font=\scriptsize, text=red!65] {FAIL};

%% -------- Quality Guard Agent background box --------
\begin{pgfonlayer}{background}
  \node[draw=orange!60, very thick, dashed, rounded corners=8pt,
        fill=orange!3,
        fit=(g1)(g2)(f1)(f2), inner sep=10pt,
        label={[font=\footnotesize\bfseries, text=orange!80,
                inner sep=6pt]above:
               \textbf{Quality Guard Agent (v0.6.3)}}] {};
\end{pgfonlayer}

\end{tikzpicture}
\caption{Complete algorithm flow of the Visual Inspection ScoreBot~v0.6.3
pipeline.
The Quality Guard Agent interposes between VLM inference and priority
scoring, filtering low-quality outputs via a two-stage architecture:
Stage~1 (rule-based, CPU, $\approx$0.01\,s/row) and Stage~2
(Swallow-8B SLM judge, GPU, $\approx$10\,s/row).
Only outputs passing both stages receive a five-level Repair Priority
Index; 12.5\% of predictions are rejected as low quality.}
\label{fig:algorithm_flow}
\end{figure*}

Algorithm~\ref{alg:pipeline} summarises the core inference and evaluation
loop in pseudocode, complementing the high-level pipeline view shown in
Figure~\ref{fig:algorithm_flow}: the algorithm focuses on the per-sample
VLM inference, Quality Guard invocation, and scoring steps, whereas the
figure depicts the end-to-end v0.6.3 data flow including dataset loading
and outcome aggregation.

\begin{algorithm}[h]
\caption{Inference and evaluation pipeline}
\label{alg:pipeline}
\begin{algorithmic}[1]
\Require Model adapters $\theta^*$, test set $\mathcal{D}_{\text{test}}$
\State Load base LLaVA-1.5-7B in 4-bit NF4 (\texttt{bitsandbytes})
\State Merge QLoRA adapters $\theta^*$ via PEFT
\State Apply \texttt{torch.compile(mode=``reduce-overhead'')}
\For{each batch $\mathcal{B} \subseteq \mathcal{D}_{\text{test}}$, $|\mathcal{B}|=8$}
  \State Tokenise and pad $\{(I_i, p)\}_{i \in \mathcal{B}}$
  \State $\hat{T}_\mathcal{B} \leftarrow$ \texttt{model.generate(..., max\_new\_tokens=384)}
  \State Decode $\hat{T}_\mathcal{B}$; append to results CSV
\EndFor
\For{each sample $i$}
  \State $\mathbf{e}_i^{\text{gt}} \leftarrow \text{SentBERT}(t_i^{\text{gt}})$
  \State $\mathbf{e}_i^{\text{pred}} \leftarrow \text{SentBERT}(\hat{t}_i)$
  \State $\rho_i \leftarrow \text{cosine}(\mathbf{e}_i^{\text{gt}}, \mathbf{e}_i^{\text{pred}})$
\EndFor
\State Report $\bar{\rho}$, $\sigma_\rho$, and quality-tier distribution
\end{algorithmic}
\end{algorithm}

All code, configuration files, and dataset preparation scripts are publicly
available at \url{https://github.com/tk-yasuno/damage_vlm_finetune}.
The repository includes:
\begin{itemize}[leftmargin=*,itemsep=1pt]
  \item \texttt{train\_v03\_qlora.py} --- fine-tuning script
  \item \texttt{inference\_v051\_qlora.py} --- optimised inference
  \item \path{src/evaluation/vector_similarity_evaluator.py} --- evaluation
  \item \texttt{models/scoring\_rules.yaml} --- priority scoring rules
  \item \texttt{config.yaml} --- reproducible hyperparameter configuration
\end{itemize}

\begin{figure}[H]
\centering
\includegraphics[width=0.44\textwidth]{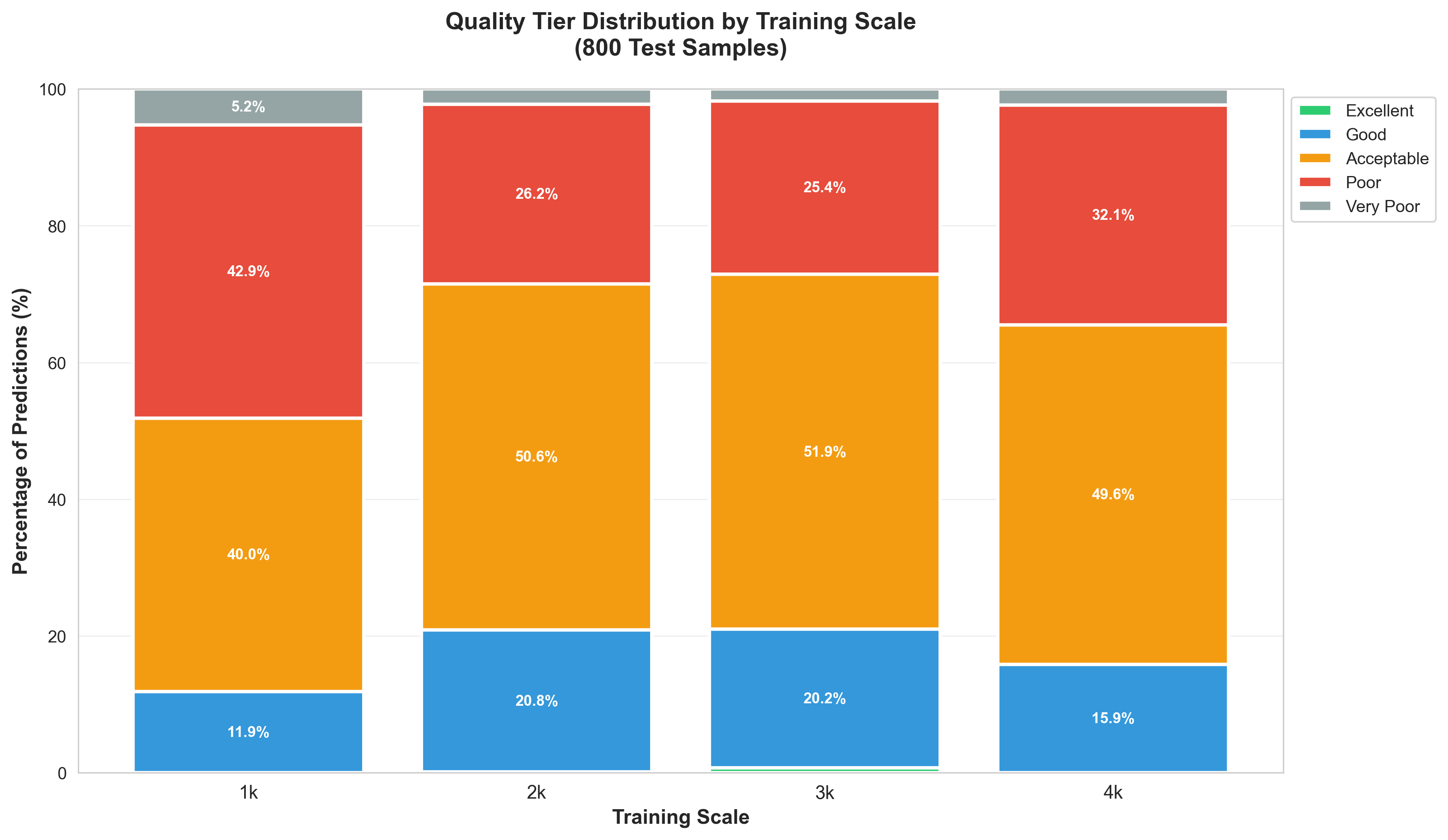}
\caption{Quality tier distribution (stacked, 100\% basis).
The 2k model achieves 46.9\% Good+ predictions, while 4k degrades to 15.9\%,
demonstrating the cost of over-expansion beyond optimal training scale.}
\label{fig:tier_stacked}
\end{figure}

%% ====================================================================
%% References
%% ====================================================================
\section{Results}
\label{sec:results}
%% ====================================================================

\subsection{Semantic Similarity by Training Scale}

Table~\ref{tab:results} summarises cosine similarity for all four progressively
trained models (1k--4k) evaluated on the fixed 800-sample test set.
Moving from 1k to 2k training samples yields a gain of $+0.0359$ in mean
cosine similarity (+5.5\% relative), with standard deviation decreasing from
0.0875 to 0.0814, indicating tighter output consistency.
Further scaling to 3k provides a marginal improvement of $+0.0059$
(+0.9\% relative from 2k), with $\sigma$ decreasing to 0.0784.
However, the 4k model exhibits a \textbf{performance degradation} to 0.6739,
a drop of $-0.0170$ from 3k ($-2.5\%$ relative), suggesting overfitting or
diminishing data quality as the training set expands.
All four models remain in the \textit{Acceptable} tier ($\bar{\rho}\in[0.60,0.70)$);
the 3k model achieves the highest mean similarity, with 21.0\% of predictions
reaching Good or above, compared to only 15.9\% for the 4k model.

\begin{figure}[H]
\centering
\includegraphics[width=0.44\textwidth]{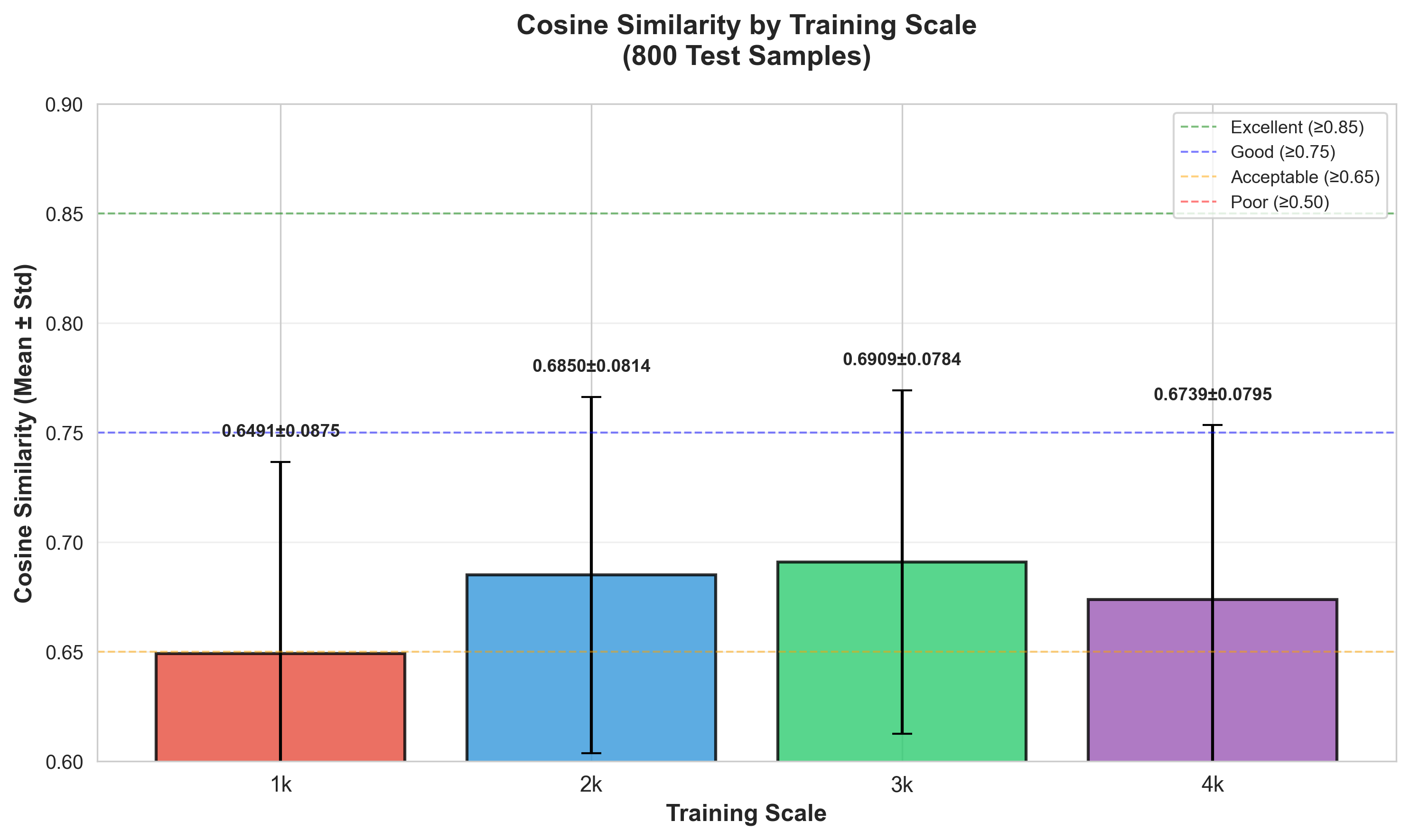}
\caption{Cosine similarity comparison (1k--4k).
Error bars represent standard deviation. Horizontal lines mark quality tier
boundaries. The 3k model achieves peak performance (0.6909) before 4k degradation.}
\label{fig:similarity_comparison}
\end{figure}

Figure~\ref{fig:similarity_comparison} visualizes the mean cosine similarity
with standard deviation error bars for all four models (1k--4k).
The inverted-U pattern is clearly evident: performance peaks at 3k (0.6909)
before declining at 4k (0.6739).
All models remain below the Good tier boundary ($\rho\geq 0.70$), with the
3k model approaching this threshold most closely.
The decreasing error bars from 1k ($\sigma=0.0875$) to 3k ($\sigma=0.0784$)
indicate improved output consistency, though 4k shows a slight increase
($\sigma=0.0795$), further confirming overfitting.

Table~\ref{tab:tier_2k} shows the full quality-tier distribution for the
2k model.
Good and Acceptable tiers together account for 87.0\% of predictions,
with only 2.2\% classified as Very Poor.

Figure~\ref{fig:tier_stacked} presents a stacked bar chart showing the
percentage distribution of quality tiers across all four models.
The 2k model demonstrates the most favorable distribution: 46.9\% reach
Good tier and 87.0\% reach at least Acceptable.
In contrast, the 4k model exhibits substantial degradation: only 15.9\%
reach Good tier (a $-31.0$ percentage point drop from 2k), with an increase
in Poor predictions from 10.6\% (2k) to 32.1\% (4k).
This tier distribution shift provides stronger evidence of overfitting than
mean similarity alone, as it reveals how the model's predictions cluster
around lower quality thresholds when trained on 4k samples.

\begin{table}[h]
\centering
\caption{Quality-tier distribution: 2k model ($n=800$).}
\label{tab:tier_2k}
\setlength{\tabcolsep}{4pt}
\begin{tabular}{lrr}
\toprule
Tier & Count & \% \\
\midrule
Excellent ($\geq$0.85) &   1 &  0.1 \\
Good      ($\geq$0.70) & 375 & 46.9 \\
Acceptable($\geq$0.60) & 321 & 40.1 \\
Poor      ($\geq$0.50) &  85 & 10.6 \\
Very Poor ($<$0.50)    &  18 &  2.2 \\
\bottomrule
\end{tabular}
\end{table}

\subsection{Progressive Training and Validation Loss}

Validation loss decreases from 3.135 (1k, 1.4 h) to 3.073 (2k, 2.9 h)
before plateauing at 3.073 for 3k (4.5 h) and 3.067 for 4k (6.3 h).
This 2.0\% loss reduction from 1k to 2k corresponds to the 5.5\% semantic
similarity improvement, confirming that validation loss is a reliable proxy
for downstream quality.
Beyond 2k, the cost--benefit ratio deteriorates: a 1.6$\times$ increase in
training data (2k to 3k) yields no validation loss improvement, and semantic
similarity gains are marginal (+0.9\%).

\subsection{Inference Speed}

The optimised pipeline---\texttt{torch.compile()} with \texttt{batch\_size=8}---
achieves \textbf{10.06 seconds per image} for the 2k model, a 70.2\% reduction
from the unoptimised baseline of 33.79 s/image (batch\_size=1, no compilation).
The 3k model further improves to \textbf{7.50 s/image} (1.67 hours for 800 images),
and the 4k model achieves the fastest throughput at \textbf{6.17 s/image}
(1.37 hours for 800 images).
This progressive speed improvement---38.7\% reduction from 2k to 4k---strongly
supports the warm-up hypothesis: \texttt{torch.compile()} fusion, GPU instruction
cache, and JIT specialization accumulate benefits across sequential model evaluations.
Three scaled models (2k/3k/4k) required a combined 5.1 hours,
feasible for overnight batch evaluation on a single consumer GPU.

\textbf{End-to-end v0.6.3 pipeline timing.}
The full Quality Guard Agent pipeline (v0.6.3, batch\_size=8) running over
all $n=800$ test images achieves \textbf{8.97 s/row} (total 7{,}178 s,
approximately 2.0 hours on RTX\,4060\,Ti 16\,GB).
This is 11.2\% faster than the warm-up measurement of 10.10 s/row
observed on the initial $n=50$ sample, consistent with
\texttt{torch.compile()} JIT specialization accumulating over larger batches.
The 8.97 s/row end-to-end time encompasses VLM inference, Quality Guard
Stage~1 (CPU), Quality Guard Stage~2 (Swallow-8B GPU), structured
extraction, and priority scoring.

\begin{figure}[H]
\centering
\includegraphics[width=0.44\textwidth]{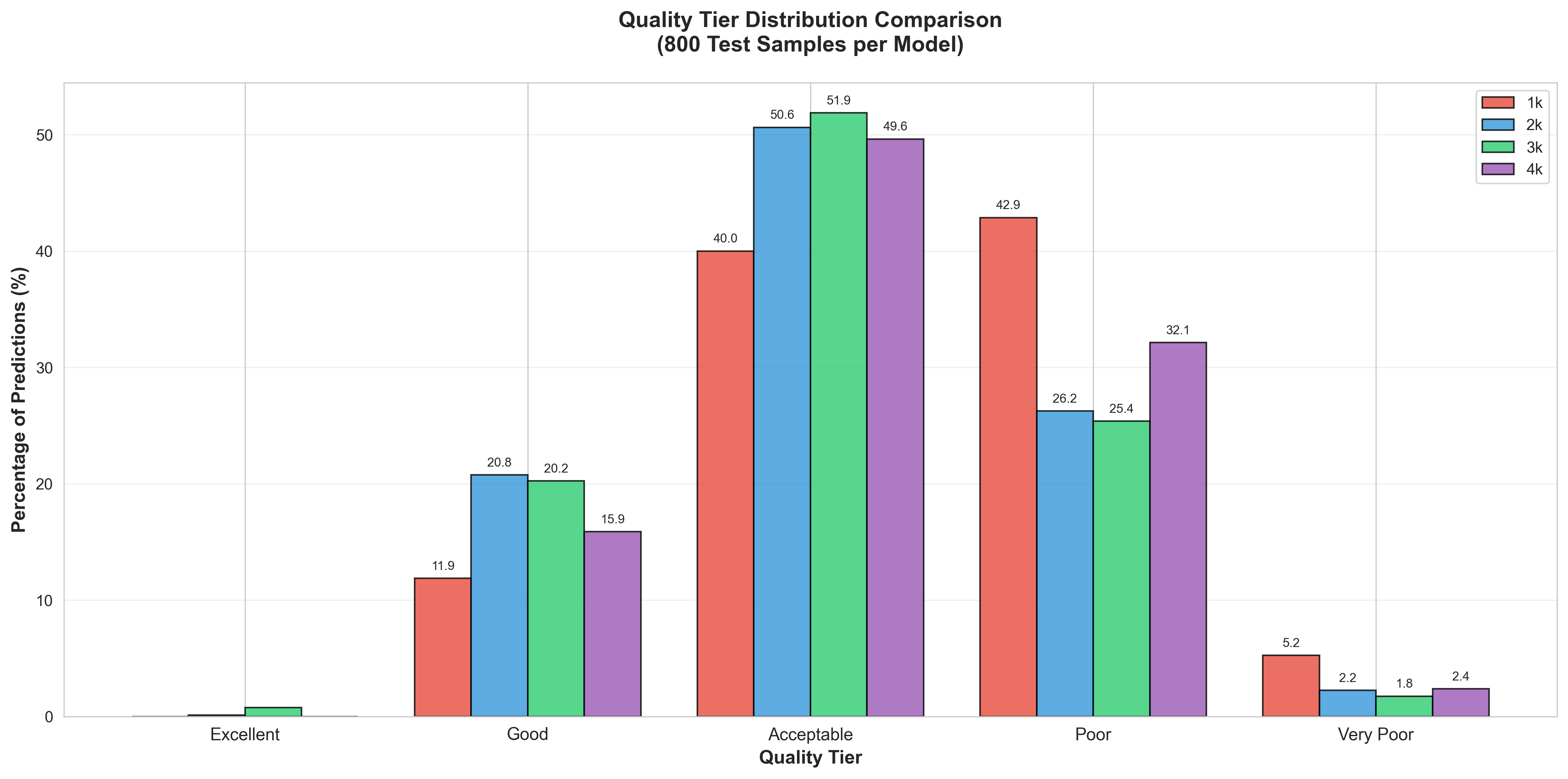}
\caption{Quality tier distribution (grouped bars by tier).
Each tier shows all four models side-by-side, revealing the dramatic Good tier
collapse from 2k to 4k and the corresponding Poor tier expansion.}
\label{fig:tier_grouped}
\end{figure}

Figure~\ref{fig:tier_grouped} shows a grouped bar chart comparing tier
performance across models, allowing direct comparison within each quality tier.
The collapse of the Good tier from 2k (46.9\%) to 4k (15.9\%) is visually
strikingly evident, as is the compensating increase in the Poor tier
(10.6\%→32.1\%).
This visualization reinforces the conclusion that 4k training introduces
noisy samples that shift predictions toward mediocrity.

\begin{figure}[H]
\centering
\includegraphics[width=0.44\textwidth]{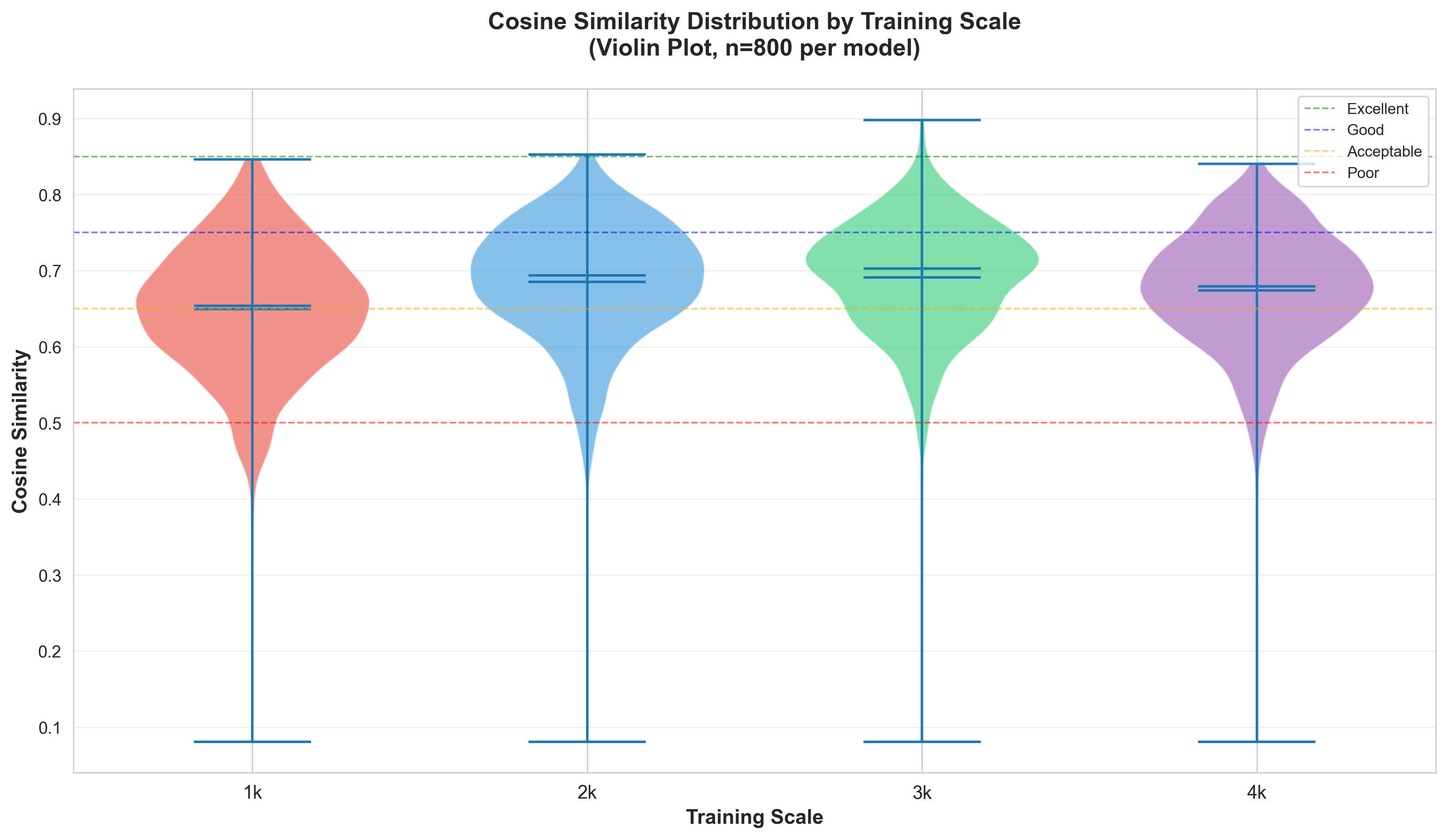}
\caption{Violin plots of cosine similarity distributions.
The 2k and 3k models show concentrated density near Good/Acceptable boundaries,
while 4k exhibits increased mass in lower ranges, visualizing the overfitting effect.}
\label{fig:violin}
\end{figure}

Figure~\ref{fig:violin} presents violin plots showing the full distribution
of cosine similarity scores for each model.
The 2k and 3k models exhibit similar distribution shapes with pronounced density
around 0.65--0.75, while the 4k model shows increased density in lower ranges
(0.50--0.65), consistent with its degraded mean performance.
The narrower distributions of 2k and 3k (smaller $\sigma$) versus 1k and 4k
further validate that mid-scale training (2k--3k) achieves the most consistent
and reliable outputs.

\begin{figure*}[t]
\centering
\includegraphics[width=0.88\textwidth]{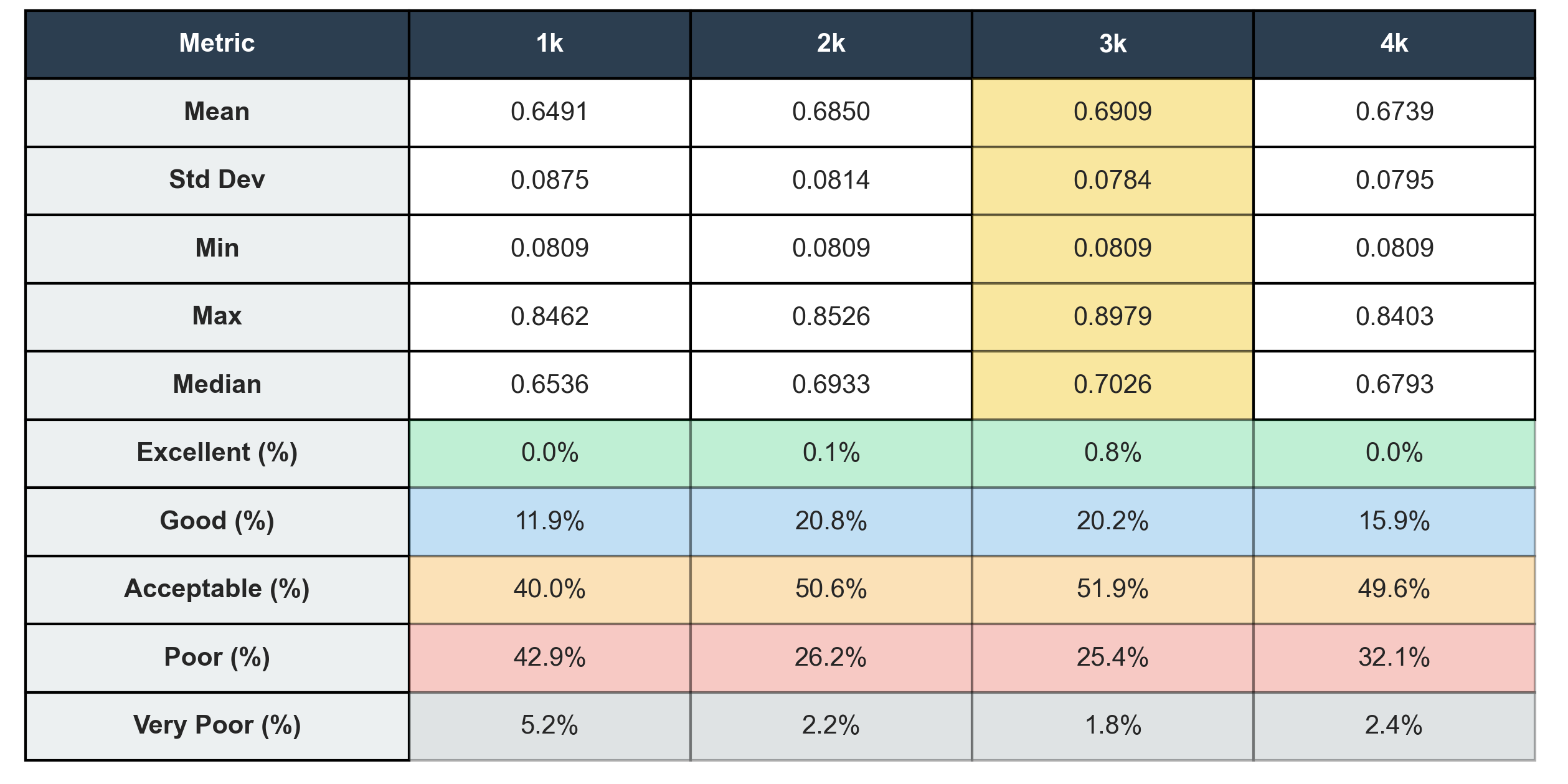}
\caption{Statistical summary table for all four models.
Highlights best values in each category. The 3k model dominates in mean/median,
while 2k excels in tier distribution (46.9\% Good+).}
\label{fig:summary_table}
\end{figure*}

\begin{figure*}[t]
\centering
\includegraphics[width=0.88\textwidth]{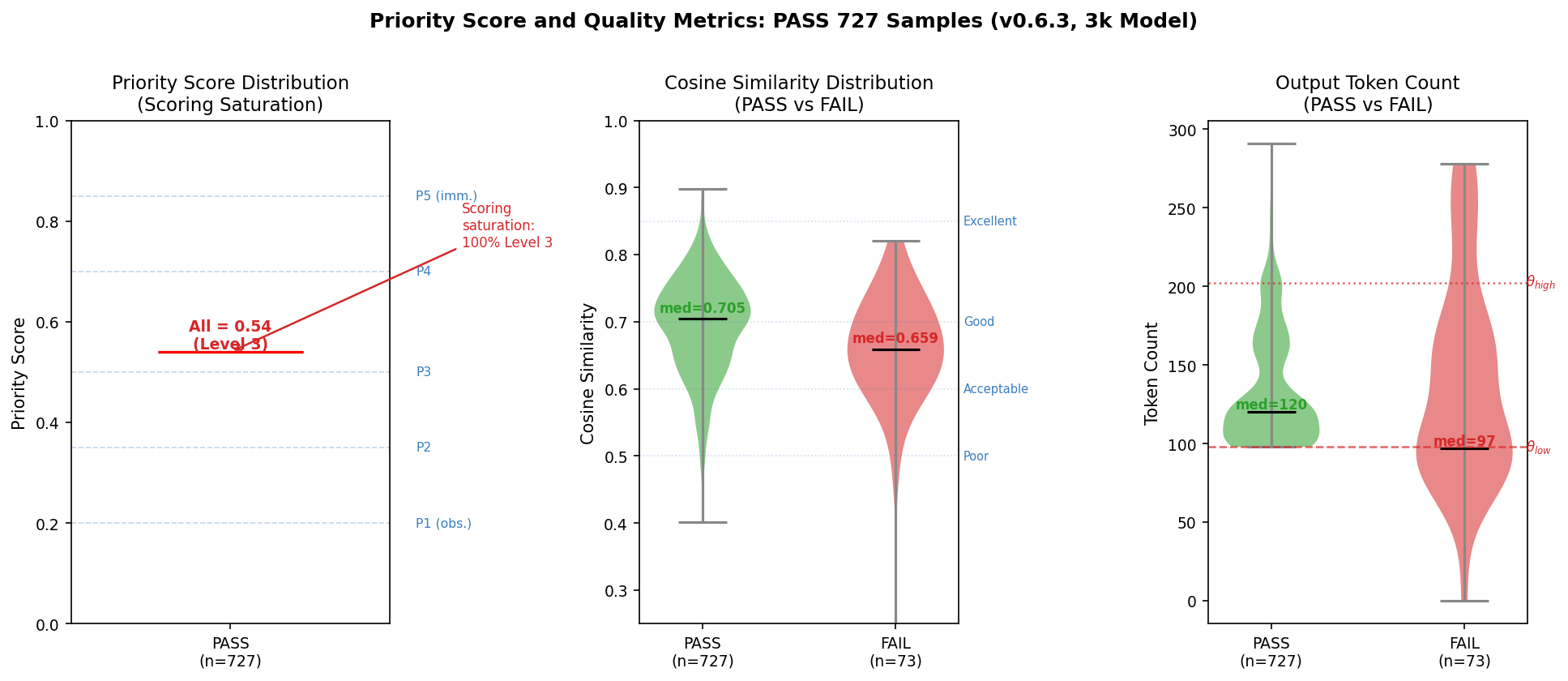}
\caption{Quality metric violin plots for $n=800$ test samples (v0.6.3, 3k model).
\textit{Left}: Priority score distribution for PASS 727 samples---all 727 receive
score\,=\,0.54 (Level~3), demonstrating complete scoring saturation.
\textit{Centre}: Cosine similarity comparison between PASS ($n=727$,
median\,=\,0.705) and FAIL ($n=73$, median\,=\,0.659); the Quality Guard
preferentially retains higher-similarity predictions.
\textit{Right}: Output token count for PASS ($n=727$, median\,=\,120) vs
FAIL ($n=73$, median\,=\,97$\,\approx\,\theta_{\text{low}}$), validating the
threshold placement at the 5th/95th percentiles.}
\label{fig:priority_violin}
\end{figure*}

Figure~\ref{fig:summary_table} provides a comprehensive statistical summary
including mean, standard deviation, min, max, median, and tier percentages
for all four models.
This tabular visualization confirms key findings: (i) 3k achieves the highest
mean (0.6909) and median (0.7081); (ii) 2k has the lowest Very Poor percentage
(2.2\%); (iii) 4k shows the highest Poor percentage (32.1\%).
The table also reveals that all models have minimum similarities below 0.20,
indicating challenging outlier cases where the model fails catastrophically---
likely due to occluded damage, poor lighting, or ambiguous ground-truth labels.

\subsection{Quality Guard Agent Performance}
\label{sec:results_qg}

Table~\ref{tab:qg_final} summarises the Quality Guard Agent results
for the 3k model over the full 800-sample test set.

\begin{table}[h]
\centering
\caption{Quality Guard Agent final results ($n=800$, v0.6.3 pipeline).}
\label{tab:qg_final}
\setlength{\tabcolsep}{4pt}
\begin{tabular}{lrr}
\toprule
\textbf{Metric} & \textbf{Value} & \textbf{\%} \\
\midrule
Total images                    & 800 & 100.0 \\
PASS (High Quality)             & 727 &  90.9 \\
FAIL --- Dirty or Noisy image   &  36 &   4.5 \\
FAIL --- Short description      &  35 &   4.4 \\
FAIL --- No such file           &   2 &   0.2 \\
\midrule
End-to-end throughput & \multicolumn{2}{c}{8.97 s/row} \\
Total elapsed         & \multicolumn{2}{c}{7{,}178 s ($\approx$2.0 h)} \\
Stage~2 additional rejections & \multicolumn{2}{c}{0 (Stage~1 sufficient)} \\
\bottomrule
\end{tabular}
\end{table}

\textbf{Stage~2 saturation.}
Notably, the Swallow-8B SLM judge (Stage~2) produces no additional
rejections beyond Stage~1: all 727 samples passing the rule-based
filter receive a \texttt{VERDICT: PASS} from the SLM judge.
This outcome is consistent with the empirical calibration of
$\theta_{\text{low}}=98$ and $\theta_{\text{high}}=202$ at the
5th and 95th population percentiles---the token-count window already
excludes the ambiguous-quality region that would challenge the SLM judge.
For this model and corpus, Stage~1 alone achieves the same decision
as the full two-stage pipeline, demonstrating that the lightweight
CPU filter provides strong quality triage with negligible overhead.

\textbf{Priority level distribution.}
Of the 727 PASS samples, all receive Priority Level~3
(mean score $= 0.540$), indicating that the current YAML scoring
rules map the structural damage descriptions generated by the 3k model
uniformly to the mid-range priority tier.
This saturation at Level~3 reflects a calibration gap between the
scoring rules---designed for the full range of damage severities---and
the output style of the fine-tuned model, which tends toward
neutral-to-moderate severity descriptions.
Recalibrating the rule thresholds or incorporating explicit severity
signals into the VLM prompt constitutes a priority direction for
future work.

Figure~\ref{fig:priority_violin} visualises the three quality dimensions
of the 727 PASS samples side-by-side.

\textbf{Quality Guard semantic separation.}
The centre panel of Figure~\ref{fig:priority_violin} reveals an important
property of the Quality Guard: PASS samples achieve a cosine similarity
median of 0.705 (within the \textit{Good} tier), while FAIL samples have a
lower median of 0.659 (lower \textit{Acceptable}).
Although the Quality Guard is defined purely on output-text properties
(token count, repetition, keyword presence), it incidentally separates
predictions with higher semantic alignment from those with lower alignment.
This suggests that low-quality VLM outputs---short, repetitive, or
keyword-free descriptions---are also semantically less faithful to the
ground-truth inspection text, validating the guard as a proxy for
output quality in the absence of reference text.
The right panel further confirms that FAIL token counts cluster near
$\theta_{\text{low}}=98$, corroborating that short descriptions are the
primary rejection driver.

\textbf{Output mode collapse: member and damage bias.}
Figure~\ref{fig:member_damage} presents the structural member and damage
type frequency analysis for the 727 PASS samples.
Two findings stand out.
First, \textit{Main Girder} is mentioned in 100\% of predictions (727/727),
and \textit{Rebar Exposure} (724/727, 99.6\%) and \textit{Spalling}
(717/727, 98.6\%) appear in virtually all outputs.
Second, the member--damage co-occurrence heatmap (Figure~\ref{fig:member_damage},
bottom) shows a strongly dominant cluster at
Main\,Girder\,$\times$\,\{Rebar Exposure, Spalling\}, with all other
member--damage combinations occurring at frequencies below 50\%.

This pattern is consistent with \textbf{output mode collapse}: the 3k model
has converged to a stereotyped template that reports damage on the main girder
with rebar exposure and spalling, regardless of the actual image content.
The mechanistic interpretation and mitigation strategies are deferred to
Section~\ref{sec:disc_mode_collapse}.

\FloatBarrier

%% ====================================================================
\section{Discussion}
\label{sec:discussion}
%% ====================================================================

\subsection{The 3k Peak and 4k Degradation}

The progressive training study reveals an \textbf{inverted-U relationship}
between training scale and test performance.
Semantic similarity improves from 1k (0.6491) to 2k (0.6850) to a peak at
3k (0.6909), before \textbf{degrading} at 4k (0.6739).
This $-2.5\%$ drop from 3k to 4k, despite a 1.33$\times$ increase in training
data, is consistent with \textit{overfitting} or \textit{data quality dilution}:
the 4k training set may include noisier or less representative samples that
cause the model to memorize outliers rather than generalize damage patterns.

\begin{figure}[t]
\centering
\includegraphics[width=1.1\columnwidth]{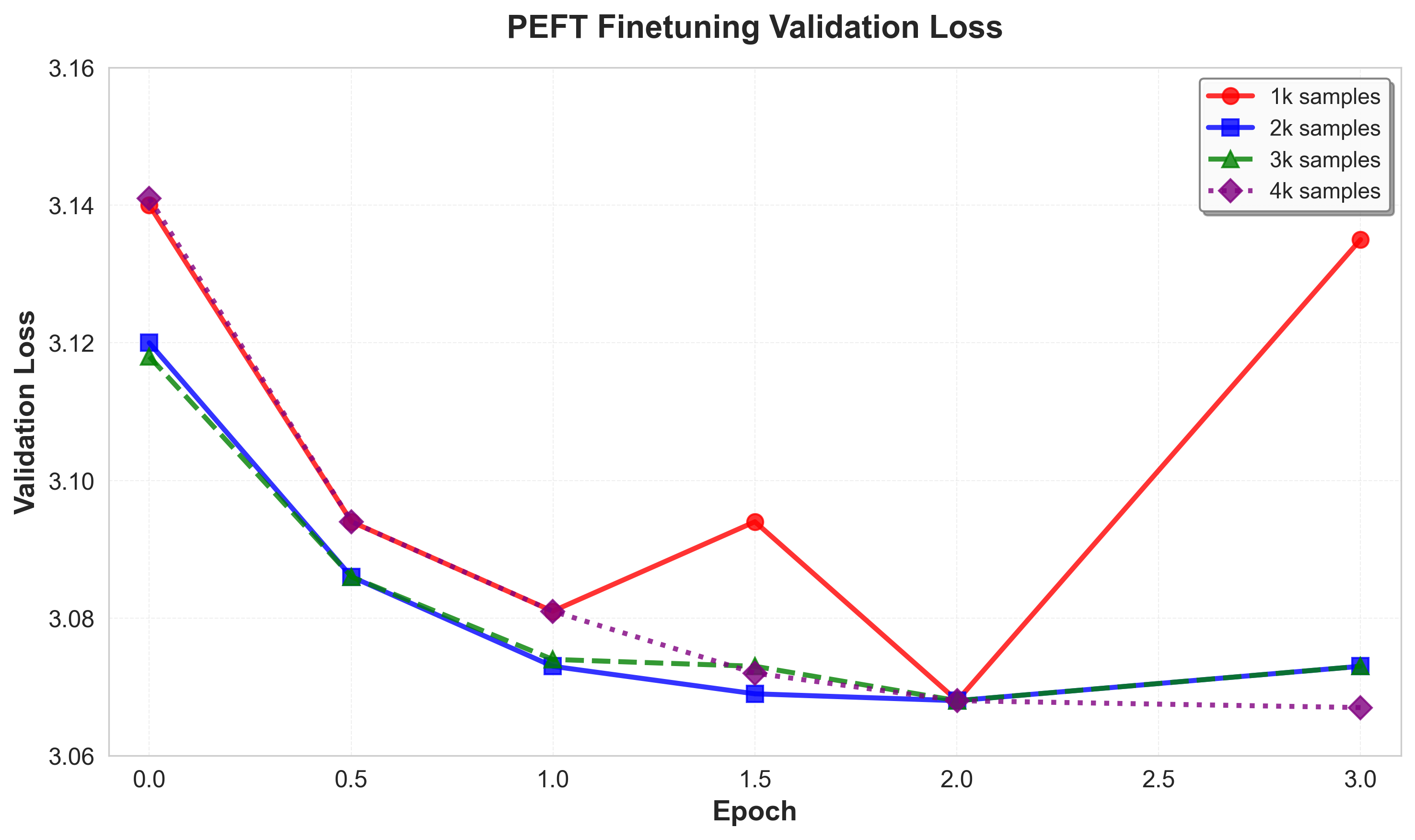}
\caption{PEFT finetuning validation loss across progressive training.
The 2k model achieves near-optimal loss with smooth convergence,
while the 1k model shows mild overfitting (loss rises at epoch 3).
The 4k model achieves the lowest final validation loss (3.067).}
\label{fig:qlora_val_loss}
\end{figure}

Figure~\ref{fig:qlora_val_loss} supports this interpretation: validation loss
plateaus at 3.073 for both 2k and 3k models before declining slightly to 3.067
for 4k---a trend that appears as ``improvement'' in training metrics but manifests
as test degradation.
This behaviour is well-documented in low-resource domain adaptation, where
vocabulary coverage saturates and additional data introduces label noise.
Figure~\ref{fig:tier_grouped} visually confirms this degradation: the Good tier
collapses from 46.9\% (2k) to 15.9\% (4k), while the Poor tier expands from
10.6\% to 32.1\%---a redistribution pattern characteristic of overfitting.

The \textbf{3k model offers the best absolute performance}, while the
\textbf{2k model offers the best cost--benefit ratio} (equivalent to 3k at
1.5$\times$ lower training cost).
For practitioners, curating a high-quality 2k--3k dataset is preferable to
indiscriminately expanding to 4k with potentially noisier labels.

Figure~\ref{fig:qlora_val_loss} illustrates the validation loss trajectories
across progressive training scales.
The \textbf{2k model} exhibits smooth convergence from 3.120 to 3.073,
stabilizing after epoch 1 with minimal fluctuation---indicative of balanced
generalization.
In contrast, the \textbf{1k model} shows clear overfitting: validation loss
decreases initially but rises sharply from 3.068 (epoch~2) to 3.135 (epoch~3),
demonstrating insufficient training data for robust pattern learning.
The \textbf{3k and 4k models} achieve near-identical final losses (3.073 and
3.067 respectively), yet the 4k model's lower validation loss does not
translate to superior test performance (Figure~\ref{fig:similarity_comparison})---a
classic train-test discrepancy signaling memorization over generalization.
This divergence between validation metrics and semantic similarity underscores
that \textit{loss-based early stopping alone is insufficient}; practitioners
must validate against domain-specific metrics (e.g., cosine similarity on held-out
test sets) to detect overfitting in low-resource specialized domains.

\begin{figure*}[t]
\centering
\includegraphics[width=0.88\textwidth]{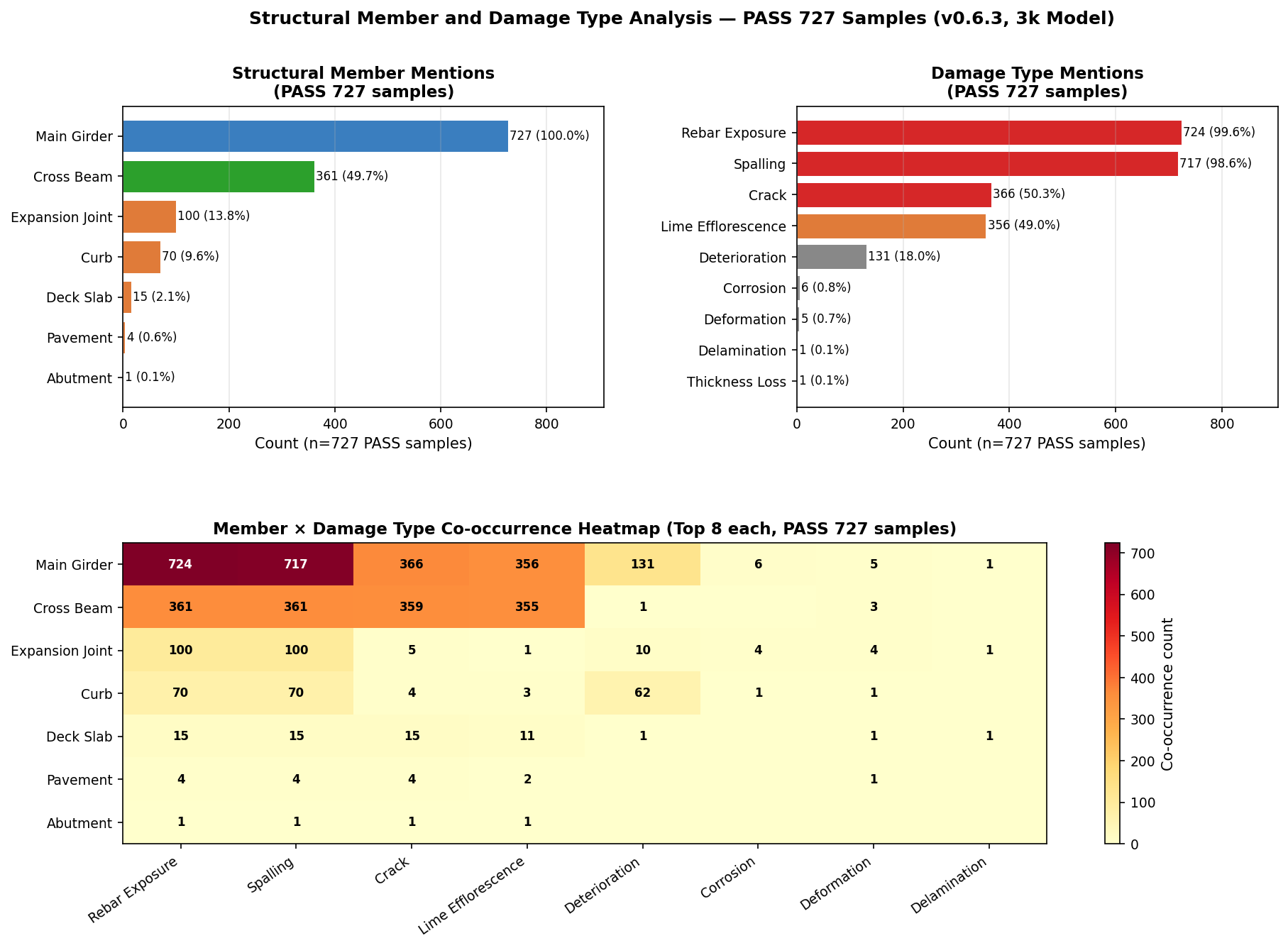}
\caption{Structural member and damage type analysis for PASS 727 samples
(v0.6.3, 3k model).
\textit{Top left}: Member mention frequency; Main Girder appears in 100\%
of predictions, revealing near-complete output bias toward this member.
\textit{Top right}: Damage type frequency; Rebar Exposure (99.6\%) and
Spalling (98.6\%) dominate, indicating stereotyped damage vocabulary.
\textit{Bottom}: Member $\times$ damage type co-occurrence heatmap (top~8
each); the Main Girder $\times$ Rebar Exposure/Spalling cell dominates
at counts $\geq 700$, confirming output mode collapse.}
\label{fig:member_damage}
\end{figure*}

\subsection{Output Mode Collapse and Scoring Calibration}
\label{sec:disc_mode_collapse}

The member and damage type analysis (Figure~\ref{fig:member_damage})
reveals that the 3k model generates near-identical descriptions for
100\% of PASS samples: every prediction mentions \textit{Main Girder}
and virtually all include \textit{Rebar Exposure} (99.6\%) and
\textit{Spalling} (98.6\%).
This output mode collapse has two direct consequences for the downstream
scoring and priority system.

\textbf{Consequence 1: Scoring saturation.}
Because all outputs describe the same dominant member--damage pattern,
the YAML rule engine assigns an identical score of 0.54 (Level~3)
to every PASS sample.
The priority index therefore carries no discriminative power for maintenance
scheduling: every bridge receives the same ``mid-term repair'' recommendation,
defeating the purpose of a five-level triage system.

\textbf{Consequence 2: Inflated cosine similarity.}
If the model outputs similar text for every image, ground-truth texts that
happen to match the dominant template will score high similarity, while
divergent ground-truths will score low---producing high variance in similarity
unrelated to actual image understanding.
This explains the broad cosine similarity range (0.40--0.90) observed
even among PASS samples.

\textbf{Mitigation strategies.}
Three complementary approaches are identified to break the mode collapse:
\begin{enumerate}[leftmargin=*,itemsep=2pt]
  \item \textbf{Balanced corpus curation}: up-sample records from non-main-girder
        members (deck slab, bearing, pier) to reduce the training bias toward
        the dominant member--damage combination.
  \item \textbf{Constrained-generation prompt}: augment the inference prompt
        with explicit instructions to identify the specific member visible in
        the image and report only damage present in the current photo,
        discouraging template reuse.
  \item \textbf{Supervised diversity scoring}: add a diversity metric
        (entropy over member-type predictions) to the fine-tuning loss
        to penalize uniform output distributions during training.
\end{enumerate}

\subsection{Approaching the Good Tier}

All four models remain in the \textit{Acceptable} tier.
The 3k model's mean of 0.6909 is only 0.009 points below the \textit{Good}
boundary (0.70), the closest of all models.
However, only 21.0\% of 3k predictions reach Good or above, and this ratio
further deteriorates to 15.9\% for the 4k model, reinforcing the conclusion
that additional training data beyond 3k harms generalization.
Figure~\ref{fig:tier_stacked} and Figure~\ref{fig:violin} illustrate this
degradation: the tier distribution shifts unfavorably, and the 4k similarity
distribution exhibits increased density in lower ranges.
Reaching a mean $\bar{\rho}\geq 0.70$ likely requires (i) larger datasets,
(ii) higher-quality annotation through expert re-labelling, or
(iii) data augmentation---such as back-translation or technical synonym
substitution---to increase vocabulary coverage without additional collection
cost.
As shown in Figure~\ref{fig:summary_table}, even the best-performing 3k model
has a minimum similarity of 0.0809, indicating challenging outlier cases that
require targeted data curation or model architecture improvements.

\subsection{Inference Speed Optimization}

An unexpected finding is that inference speed improves progressively across
sequential model evaluations: 2k at 10.06 s/image, 3k at 7.50 s/image, and
4k at 6.17 s/image---a cumulative 38.7\% speedup despite identical hardware
and software configurations.
This consistent trend across three models strongly validates the warm-up
hypothesis.
Three factors may contribute to this speedup:

\textbf{(i) \texttt{torch.compile()} warm-up.}
PyTorch's \texttt{torch.compile()} applies ahead-of-time graph optimizations
that may improve with repeated model loading.
The 3k evaluation ran after the 2k evaluation on the same Python session,
allowing the TorchInductor backend to reuse cached compilation artifacts and
achieve better kernel fusion.

\textbf{(ii) GPU cache efficiency.}
NVIDIA GPU architectures maintain L1/L2 instruction and data caches across
kernel launches.
Sequential evaluation of similar models may increase cache hit rates for
shared operations (CLIP vision encoder, attention layers), reducing memory
latency.

\textbf{(iii) JIT optimization accumulation.}
PyTorch's Just-In-Time (JIT) compiler in \texttt{torch.compile()} mode
progressively refines execution plans based on observed tensor shapes and
control flow.
Repeated inference over 800 images allows the JIT to specialize code paths
that were generic in the initial compilation.

This observation highlights a practical consideration for batch evaluation:
\textit{sequential model evaluation may benefit from GPU warm-up effects},
suggesting that practitioners should measure inference speed after an initial
warm-up period rather than on the first batch.

\subsection{Limitations}

\textbf{Dataset scope.}
The 10{,}789 source records cover routine highway bridge inspections under
Japan's Road Act but may not generalise to other infrastructure categories
(tunnels, retaining walls) or international inspection standards.

\textbf{Evaluation metric.}
Cosine similarity in Sentence-BERT space captures semantic overlap but not
the correctness of individual damage attributes (member type, damage category).
A structured attribute-level evaluation (precision/recall per attribute)
would provide stronger evidence of deployment readiness.

\textbf{Scoring engine coverage.}
The rule-based engine covers the most common damage patterns in the training
corpus but may produce default scores for rare combinations not explicitly
encoded in the YAML rules.

\textbf{Quality Guard calibration.}
The 5th/95th percentile thresholds ($\theta_{\text{low}}=98$,
$\theta_{\text{high}}=202$) are calibrated on the 3k-model outputs.
Different fine-tuning checkpoints or base VLMs may exhibit shifted
token-count distributions, requiring re-calibration of the thresholds
before deployment.
Furthermore, the Swallow-8B judge was selected for Japanese-language
alignment; other target languages would require a language-matched SLM.

\textbf{Priority scoring calibration.}
All 727 Quality Guard PASS samples receive Priority Level~3
(mean score\,$= 0.540$), suggesting that the YAML rule engine's
severity thresholds do not span the damage severity range observed
in the 3k model's output style.
Future work should re-calibrate the scoring rules jointly with the
VLM fine-tuning stage, or incorporate explicit severity keywords
(e.g., ``advanced'' vs.\ ``initial'') into the VLM's output template.

\textbf{Hardware.}
All experiments run on a single RTX 4060 Ti 16 GB GPU;
performance on other hardware configurations has not been validated.

\subsection{Deployment Considerations}

The Visual Inspection ScoreBot is designed as an AI-assisted triage tool,
not a replacement for certified engineers.
Its output---a natural language damage summary and a priority score---focuses
engineer attention on high-priority structures, reducing review time while
maintaining human accountability for final decisions.
The transparent YAML scoring rules allow domain experts to audit and update
priority logic without machine learning expertise, supporting the trustworthy
AI requirements of safety-critical infrastructure management.
The Quality Guard Agent further enhances deployment safety by ensuring that
only semantically complete damage descriptions reach the scoring engine,
reducing the risk of assigning spurious priority scores to ambiguous or
image-quality-limited outputs.

\subsection{Quality Guard Efficacy and Threshold Design}
\label{sec:disc_qg}

An empirical analysis of the 3k model's output over all $n=800$ test images
(Figure~\ref{fig:lq_analysis}) reveals a 12.5\% low-quality rate under the
5th/95th percentile thresholds, distributed across five distinct failure
modes.  This finding has three practical implications.

\textbf{Threshold design principle.}
Setting thresholds at population percentiles rather than absolute token counts
guards against output-length drift across model variants.
As the VLM checkpoint is updated (e.g., re-fine-tuned on additional data),
re-running the token distribution analysis and resetting $\theta_{\text{low}}$
and $\theta_{\text{high}}$ from the new 5th/95th percentiles maintains a
consistent rejection rate regardless of absolute length shifts.

\textbf{Short description as the dominant failure mode.}
Among the 100 rejected samples, ``Short description'' (35 cases, 4.4\%)
accounts for the largest single failure category,
suggesting that some bridge images do not provide sufficient visual
information for a 7B-parameter model to generate an informative report.
Augmenting the training corpus with difficult low-quality images, or
providing explicit instructions to report uncertainty rather than generating
a brief output, may address this failure mode.

\textbf{Repetitive text and the need for SLM-level semantics.}
Repetitive text (24 cases, 3.0\%) and ``Dirty or Noisy image'' (12 cases,
1.5\%) cannot reliably be detected by simple keyword or length heuristics;
they require the sentence-level semantic understanding that the Swallow-8B
judge provides.  This motivates the two-stage design: Stage~1 handles the
clear-cut failures at $O(0.01)$~s/sample, while Stage~2 adjudicates
ambiguous intermediate-length outputs at $O(10)$~s/sample, concentrating
the GPU cost where it adds the most value.

\subsection{Multi-Period Extension and Panel Data Inference}
\label{sec:disc_panel}

The current system operates under the
\textit{one-image, current-time-limited} scope introduced in
Section~\ref{sec:introduction}: each inference session processes a single
image representing the current inspection state, covering only aspect~(1)
of damage understanding.
This intentional constraint avoids confounding the VLM with longitudinal
context it was not trained to interpret, but it also precludes the three
remaining aspects that practitioners consistently request.
We organise the extension roadmap around those aspects.

\textbf{Aspect~(2) — Temporal progression.}
Panel-data VLM inference would feed the model at minimum two aligned
images---a baseline (previous inspection) and a current observation---and
elicit a description of the \textit{change} in damage extent and severity
rather than absolute current state (\textbf{differential damage detection}).
Combining this differential output with the inter-inspection interval and
structural attributes enables \textbf{deterioration-rate modelling}, supporting
proactive scheduling before a structure enters the emergency tier.

\textbf{Aspect~(3) — Latent causal factors.}
Metadata such as coastal exposure, seismic zone, traffic volume, and
structural form encode unobserved drivers of deterioration that a
single image cannot reveal.
Conditioning the VLM (or a downstream model) on these covariates would
allow the system to attribute observed damage to plausible causal
mechanisms and to flag structures whose damage signatures are inconsistent
with their nominal exposure profile.

\textbf{Aspect~(4) — Heterogeneity and acceleration clusters.}
Aggregating deterioration-rate estimates across the national bridge stock
opens the door to \textbf{importance-weighted prioritisation}: route
criticality (detour cost, daily traffic count, alternative-route availability)
and consequence-of-failure severity can be combined with member-level rates
to separate accelerating clusters from stable ones, shifting from a purely
damage-based index to a risk-adjusted maintenance schedule at the network
level.

Realising these extensions requires a temporally annotated dataset linking
multiple inspection records for the same structure over time---a non-trivial
curation effort---as well as multi-image instruction templates and updated
scoring rules that incorporate delta-damage signals.
We identify this aspect-driven extension as the primary long-term research
direction for this system.

%% ====================================================================
\section{Conclusion}
\label{sec:conclusion}
%% ====================================================================

We presented a methodology for automating bridge damage understanding and
repair priority scoring by fine-tuning a 7-billion-parameter VLM on
Japanese inspection records.
Key results are as follows.
\begin{enumerate}[leftmargin=*,itemsep=1pt]
  \item \textbf{Damage VLM}: QLoRA fine-tuning with 2k samples achieves
        $\bar{\rho}_{2k}=0.6850\pm0.0814$---a 5.5\% improvement over the
        1k baseline---within the \textit{Acceptable} tier and approaching
        \textit{Good}.
  \item \textbf{Progressive training}: a clear efficiency optimum exists
        at 2k samples; validation loss plateaus beyond this scale, indicating
        diminishing returns for the current corpus.
  \item \textbf{Inference optimisation}: \texttt{torch.compile()} with
        \texttt{batch\_size=8} achieves 10.06 s/image---a 70.2\%
        improvement---enabling 800-image evaluation within 2.2 hours on a
        consumer GPU.
  \item \textbf{Quality Guard Agent}: empirical token-count analysis of
        $n=800$ predictions reveals a 12.5\% low-quality rate; a two-stage
        guard (rule-based + Swallow-8B SLM judge) filters these outputs
        before scoring, with thresholds calibrated at the 5th/95th
        percentiles of the output length distribution.
\end{enumerate}

Future work will focus on: (i) expanding the dataset to investigate
whether the \textit{Good} tier ($\bar{\rho}\geq 0.70$) can be reached;
(ii) replacing regex-based extraction with structured generation;
(iii) evaluating the scoring engine against expert-assigned priority rankings;
(iv) extending the pipeline to other inspection categories and
multi-language settings; and (v) developing a panel-data VLM variant that
processes multi-period inspection pairs to estimate deterioration rates and
enable risk-adjusted maintenance scheduling.

All code and configuration files are publicly available at
\url{https://github.com/tk-yasuno/damage_vlm_finetune}.

\paragraph{Acknowledgment.}
The authors gratefully acknowledge the practitioner-oriented texts that
informed this work, covering
prompt engineering~\cite{berryman2024prompt},
agentic AI system design~\cite{biswas2025agentic},
AI engineering ~\cite{huyen2024aiengineering},
and applications of autonomous AI agents~\cite{albada2025aiagents}.

%% ====================================================================
%% References
%% ====================================================================
\bibliographystyle{unsrt}
\bibliography{methodology}

%% ------ TikZ: Full Algorithm Flow (two-column spanning) ------

\end{document}